\documentclass[letterpaper]{article} 
\usepackage{aaai2026}  
\usepackage{times}  
\usepackage{helvet}  
\usepackage{courier}  
\usepackage[hyphens]{url}  
\usepackage{graphicx} 
\usepackage{natbib}  
\usepackage{caption} 
\usepackage{algorithm}
\usepackage{algorithmicx}
\usepackage{algpseudocode}
\usepackage{newfloat}
\usepackage{listings}
\usepackage{booktabs}
\usepackage{siunitx}
\DeclareCaptionStyle{ruled}{labelfont=normalfont,labelsep=colon,strut=off} 
\floatstyle{ruled}
\newfloat{listing}{tb}{lst}{}
\floatname{listing}{Listing}
\usepackage{mathtools} 
\usepackage{appendix}

\title{Communication-Efficient Federated Learning \\ with Adaptive Number of Participants}

\title{Communication-Efficient Federated Learning \\ with Adaptive Number of Participants}
\author{
    Sergey Skorik\textsuperscript{\dag},
    Vladislav Dorofeev\textsuperscript{},
    Gleb Molodtsov\textsuperscript{},
    Aram Avetisyan\textsuperscript{},
    Dmitry Bylinkin\textsuperscript{},
    Daniil Medyakov\textsuperscript{},
    Aleksandr Beznosikov\textsuperscript{}.
}
\affiliations{
    \textsuperscript{\dag} Corresponding Author. E-mail: skorik.sn99@gmail.com
}

\usepackage{bibentry}

\newtheorem{definition}{Definition}
\newtheorem{example}{Example}
\newtheorem{problem}{Problem}

\usepackage[utf8]{inputenc}
\usepackage{multirow}
\usepackage{makecell}
\usepackage{threeparttable}    
\usepackage{amsmath,amssymb}
\usepackage{amsfonts}
\usepackage{bm}
\usepackage{algpseudocode}
\usepackage{lipsum}
\usepackage{enumitem}
\usepackage{tabularx}

\newcommand{\cc}[1]{\mathcal{#1}}

\newcommand{\bE}{\mathbb{E}}
\newcommand{\x}{\bm{x}}
\newcommand{\z}{\bm{z}}

\begin{document}

\maketitle

\begin{abstract}
Rapid scaling of deep learning models has enabled performance gains across domains, yet it introduced several challenges. Federated Learning (FL) has emerged as a promising framework to address these concerns by enabling decentralized training. Nevertheless, communication efficiency remains a key bottleneck in FL, particularly under heterogeneous and dynamic client participation. Existing methods, such as FedAvg and FedProx, or other approaches, including client selection strategies, attempt to mitigate communication costs. However, the problem of choosing the number of clients in a training round remains extremely underexplored. We introduce Intelligent Selection of Participants (ISP), an adaptive mechanism that dynamically determines the optimal number of clients per round to enhance communication efficiency without compromising model accuracy. We validate the effectiveness of ISP across diverse setups, including vision transformers, real-world ECG classification, and training with gradient compression. Our results show consistent communication savings of up to 30\% without losing the final quality. Applying ISP to different real-world ECG classification setups highlighted the selection of the number of clients as a separate task of federated learning.

\end{abstract}
\begin{links}
  \link{Code}{https://anonymous.4open.science/r/ISPFL/}
\end{links}


\section{Introduction}

The progress of machine learning has led to increasingly large models and datasets, demanding more computational resources and access to diverse data \citep{alzubaidi2021review}. To address these scaling challenges, distributed training has become a core component of modern ML pipelines. Among distributed strategies, federated learning (FL) has emerged as a key paradigm \citep{kairouz2021advances, konevcny2016federated, li2020federated_research}. It enables multiple decentralized data owners to collaboratively train models without directly sharing their raw data.
Despite its numerous promises across a wide range of tasks \citep{smith2017federated, mcmahan2017communication, verbraeken2020survey}, this paradigm presents unique challenges stemming from real-world deployment scenarios \citep{kairouz2021advances, wen2023survey}. 
Client data are typically non-identically distributed and unbalanced, with each user's data reflecting personalized usage patterns. 

Among these challenges, communication efficiency emerges as a central bottleneck. Model updates must be transmitted over constrained and often unreliable uplink bandwidth. As observed in \citep{mcmahan2017communication}, many FL systems aim to mitigate this by increasing local computation or leveraging greater parallelism. In addition to local updates, a common strategy to address communication delays is partial participation, avoiding excessive wait times for slow clients \citep{li2020federated}. Varying the number of clients meets a critical trade-off. Involving more clients accelerates convergence through broader gradient diversity, but inflates per-round communication costs. This may overwhelm limited resources or lead to participant dropout. Conversely, engaging too few clients slows learning and undermines model generalization.

While significant attention has been devoted to selecting which clients to involve, the question of how many clients to select per round remains largely underexplored, despite its impact on the scalability of FL. As existing FL methods (e.g. \textsc{FedAvg} \citep{mcmahan2017communication} and \textsc{FedProx}  \citep{li2020federated}) typically assume a fixed number of participating clients per round, our core insight is that the participant count can and should be regarded as a dynamic variable to balance communication costs and convergence speed.
We propose the \textit{Intelligent Selection of Participants} (\textsc{ISP}) approach, which dynamically chooses the number of clients sufficient to make an impact in the current round.

\section{Related Work}\label{sec:related_work}

\paragraph{Classical federated algorithms.}Scaling of model architectures and training data has stimulated the adaptation of optimization algorithms to distributed setups. Built on the foundations of distributed synchronous SGD \citep{chen2017revisitingdistributedsynchronoussgd}, the seminal Federated Averaging (\textsc{FedAvg}) algorithm \citep{mcmahan2017communication} alternates local client training with global model averaging to trade-off computation and communication. Since its introduction, the field witnessed successive refinements targeting data and system heterogeneity. To mitigate client drift under non‑IID data, \textsc{FedProx} \citep{li2020federated} added a proximal term constraining local updates, preventing excessive deviation from the global model. On this foundation, \textsc{SCAFFOLD} \citep{karimireddy2020scaffold} took a more direct approach by employing control variates to explicitly manage the drift between client and server updates. Complementary advances include \textsc{FedNova}’s normalization of update magnitudes across variable local steps \citep{wang2020tackling}. Thus, they ensured fair aggregation regardless of computational differences between clients. Furthermore, \citep{reddi2020adaptive} applied adaptive Adam-like optimization to enhance efficiency. Comprehensive surveys like \citep{li2020federated_research} covered a wide range of algorithms, showing the intensive development of this direction. 
While these advances have narrowed the performance gap to centralized training, they still rely on synchronized parameter exchanges between all clients and the server.

\paragraph{Clients sampling.}Communication efficiency in federated learning can be significantly improved by carefully selecting participating clients. Early approaches had employed static policies based on statistical contribution \citep{cho2020client, lai2021oort} or system utility \citep{nishio2019client, xu2020client, ribero2022federated}. Instead of uniform sampling, client selection based on local loss values was proposed \citep{goetz2019active}, which required prior communication. To mitigate this, \textsc{Power-of-Choice} \citep{cho2020client} sampled a subset of clients for loss evaluation. \textsc{FedCor} \citep{tang2022fedcor} further reduced communication by approximating loss changes using a Gaussian process. Statistical techniques like Importance Sampling (IS) \cite{rizk2022federated} ranked clients by variance reduction, while \textsc{DELTA} \citep{wang2023delta} advocated for selecting clients with diverse gradients. \textsc{Fed-CBS} \citep{zhang2023fed} sorted clients based on label distribution uniformity. Hybrid strategies like \textsc{Oort} \citep{lai2021oort} and \textsc{PyramidFL} \citep{li2022pyramidfl} integrated statistical and system metrics, often assuming a fixed number of clients per round. As we can see, the issue of selecting specific clients has been well studied. However, these works rely on a fixed number of selected clients, albeit on their importance. Only a small number of works concern the selection of the number of clients.

\paragraph{Number of clients choice.}Recent work has begun to explore dynamic control over the number of participating clients in a communication round. The authors of \citep{de2024adaptive} proposed an exponential decay schedule to gradually reduce the number of clients. This approach can negatively affect convergence in cases of statistical heterogeneity, since the model becomes increasingly sensitive to local shifts in client updates as it approaches the optimum. Other studies, such as \citep{chen2022fedtune}, examined how the number of clients influences fine-tuning of large transformer models in federated settings. However, this line of work does not assess dynamic schemes for selecting the number of clients and is aimed exclusively at a specific scenario of transformer fine-tuning. Finally, the work \citep{chen2021dynamic} specifically addressed the choice of the number of clients in the training round and is motivated by similar considerations. However, the proposed \textsc{AdaFL} technique is reduced to a simple linear schedule of the number of clients, which, unlike \citep{de2024adaptive}, exhibited a positive trend. 

Despite these efforts, no existing research provides a solution that determines how many clients should participate in each round based on real-time metrics such as training dynamics, data heterogeneity, and resource availability. This gap highlights the need for systematic methods that can coordinate client selection. In this paper, we take a first step in this direction.

\paragraph{Main contributions.} We summarize the key contributions of this work as follows:

\begin{itemize}
    \item \textbf{New procedure.} We propose \textsc{ISP} (Intelligent Selection of Participants), a method that dynamically determines the number of participating clients in federated learning rounds. \textsc{ISP} balances communication cost and convergence by selecting the minimal client subset that ensures model improvement.
    
    \item \textbf{Versatility.} We formulate the client count selection as an optimization problem and present a universal procedure (Algorithm~\ref{alg:ISP}). 
    This technique is adaptable to various algorithms, such as \textsc{FedAvg} \citep{mcmahan2017communication} or \textsc{SCAFFOLD}.
    
    \item \textbf{Experiments.} We conduct extensive experiments on diverse benchmarks, including synthetic \textsc{CIFAR-10} \citep{krizhevsky2009learning} and Tiny-\textsc{ImageNet} \citep{le2015tiny} datasets, and a real-world electrocardiogram (ECG) classification task. \textsc{ISP} achieves up to a 30\% reduction in communication rounds over already efficient client-sampling and gradient-compression techniques while maintaining or improving accuracy. We also perform an ablation study to dissect the contribution of each component (see Appendix \ref{app:abl}).
\end{itemize}

\section{Method}

\subsection{Problem Setup}

We consider the global training objective as the solution to
\begin{equation}\label{eq:main_obj}
    \min_{\x \in \mathbb{R}^d} \; f(\x) := \sum_{i=1}^M w_i\,f_i(\x),
\end{equation}
where  $f_i(\x) = \mathbb{E}_{\z_i\sim\mathcal{D}_i}[f_i(\x,\z_i)]$ is the expected loss of client $i$ under distribution $\mathcal{D}_i$, $\x$ are the model parameters, and $w_i$ is the client weight. We write \eqref{eq:main_obj} in weighted form to capture non-uniform sampling: e.g., in \textsc{FedAvg}, $w_i = n_i/n$, with $n_i$ local samples and $n=\sum_i n_i$ total, in \textsc{Oort} $w_i = p_i$ reflect the overall (statistical and system) utility score. 
To systematically analyze such sampling behavior in FL, we introduce the concept of a client sampling strategy, formalized below. This will allow us to unify a broad class of methods under a common mathematical structure.

\begin{definition}[Client Sampling Strategy]
A client \newline sampling strategy $\mathcal{S}_{\tau}$ at round $\tau$ is a (possibly randomized) procedure that returns a subset
\[
    C_{\tau+1} \sim \mathcal{S}_\tau, \quad
    C_{\tau+1} = \{i_1,\dots,i_{m_{\tau+1}}\} \subseteq \{1,\dots,M\},
\]
where $m_{\tau+1}=|C_{\tau+1}|$ is the number of selected clients. 
\end{definition}
This formalism provides a basis for describing and comparing different sampling mechanisms, whether uniform, deterministic, or adaptive.

\begin{example}[Uniform Sampling]
In the uniform sampling strategy, we randomly select clients without replacement:
\[
    C_{\tau+1} \sim \text{Unif}\left(\{\text{all subsets of size } m\}\right).
\]
\end{example}
\noindent This ensures unbiased gradient estimation, but does not adapt to variations in client data, compute resources, or connectivity.

\begin{example}[Active FL]
    In Active FL \citep{goetz2019active} sampling procedure, each client $i_{k_\tau}$ at round $\tau$ is sampled proportionally to its valuation $p_k$:
\[
    C_{\tau+1} = \{i_{k_\tau} \mid i_{k_\tau} = 1\}, \quad i_{k_\tau} \sim Be(p_k),
\]
where $p_k$ proportional to the reduction of the client's loss.
\end{example}
\noindent Valuation $p_k$ represents the adjacency from client to server. It can also be set based on other utilities, such as the importance of the dataset. 



\subsection{Our methodology}\label{subsec:our_methodology}
In this section, we formalize the standard federated learning setup (see Algorithm \ref{alg:fed_periodic}) that serves as the foundation for our later developments. A central server orchestrates the training over $M$ clients, each holding its own private dataset. The training proceeds in a sequence of communication \emph{rounds}. At the start of the round $\tau$, the server samples a subset of clients
\[
C_{\tau+1} \;\sim\; \mathcal{S}_\tau,\qquad |C_{\tau+1}| = m_{\tau+1},
\]
according to a specified sampling strategy $\mathcal{S}_\tau$. Only these $m_{\tau+1}$ clients participate at the current round; all others remain idle. Each active client $i\in C_{\tau+1}$ initializes from the current global model $\x^\tau$ and performs a fixed number of local gradient descent steps on its own loss $f_i$. The server can instruct the client procedure: make adjustments to the $f_i$, require special results, etc. After local training, it collects the updated parameters from each client in $C_{\tau+1}$ and aggregates them. The full training proceeds for $T$ rounds, yielding the final model~$\x^T$.

In the classical FL paradigm, the number of participating clients~$m_{\tau+1}$ remains constant across rounds. While this simplifies the analysis, it overlooks an important degree of freedom: varying $m_{\tau+1}$ can trade off communication cost against convergence speed. We can study this trade‑off by introducing the \textit{next‑round loss changing} 
\begin{equation}\label{eq:cs_obj}
\delta f_{\tau}(m_{\tau+1}) = f\left(\x^{\tau + 1}(m_{\tau+1})\right)-f(\x^\tau),
\end{equation}

\begin{algorithm}[ht]
  \caption{Periodic Communication with Fixed Client Set}
  \label{alg:fed_periodic}
  \begin{algorithmic}[1]
    \Require initial model $\x^0$, sampling strategies $\{\mathcal{S}_{\tau}\}$.
    \For{$\tau = 0$ to $T-1$}
      \State \textcolor{blue}{Define $m_{\tau+1}$} \label{line:dynamic_number} \Comment{Algorithm \ref{alg:number_strategy}}
      \State Sample clients $C_{\tau+1}\sim\mathcal{S}_\tau$ with $|C_{\tau+1}| = m_{\tau+1}$
      \State $\{\x_i^\tau\}_{i \in C_{\tau+1}} \gets \textsc{GetUpdates}(\x^\tau, C_{\tau+1})$
      \State $\x^{\tau+1} \gets Aggregate(\{\x_i^\tau\}_{i \in C_{\tau+1}})$ \label{line:aggregate}
    \EndFor
    \State \textbf{Output:} $\x^T$
    \Statex
    \Function{GetUpdates}{$\x^\tau, C$} \label{func:get_update}
      \ForAll{clients $i \in C$ \textbf{in parallel}}
        \State $\x_i^\tau \gets ClientUpdate(\x^\tau, f_i)$
      \EndFor
       \State \Return $\{\x_i^\tau\}_{i \in C_{\tau+1}}$
    \EndFunction
  \end{algorithmic}
\end{algorithm}

where $\x^{\tau + 1}(m_{\tau+1})$ and $\x^{\tau}$ are modeled according to random subsets $C_{\tau+1}\sim\mathcal{S}_\tau, C_\tau\sim\mathcal{S}_{\tau-1}$ of sizes $m_{\tau+1}, m_\tau$, respectively (see Line \ref{line:aggregate}, Algorithm \ref{alg:fed_periodic}).
The choice of $m_{\tau+1}$ governs the 
magnitude and direction of $\delta f_\tau$: larger values typically lead to more pronounced improvements in average loss, while smaller ones may even result in degradation due to insufficient update quality.

We address this trade-off by \emph{minimizing} communication costs. Then, our goal is to choose the smallest $m_{\tau+1}$ that still ensures model improvement, i.e.,
\begin{equation}\label{eq:select_m}
\begin{aligned}
  &\min_{m_{\tau+1}} \quad m_{\tau+1},\\
  &\text{s.t.}\quad \bE_{C_{\tau+1}} \delta f_\tau(m_{\tau+1}) < 0.
\end{aligned}
\end{equation}
Note that $\delta f_\tau(m_{\tau+1})$ also depends on random $C_{\tau+1}$, $\textstyle \x^{\tau + 1}(m_{\tau+1}) = \x^{\tau + 1}(m_{\tau+1}, C_{\tau+1})$. Hence, the model improvement is implied as averaged. Solving \eqref{eq:select_m} provides an adaptive client selection mechanism, ensuring minimally consumed resources while provably maintaining the next-round model improvement. By dynamically balancing $m_{\tau+1}$ against the expected loss reduction, it reconciles efficiency and convergence in stochastic federated environments. We also explore an alternative optimization formulation (see Appendix \ref{app:abl:optim_problem}). However, while the problem formulation is complete, several critical challenges currently prevent its direct solution. We now systematically examine these limitations and propose corresponding resolution strategies.

\begin{problem}[Direct evaluation]\label{problem:direct_evaluation}
    The direct computation of $\delta f_\tau(m_{\tau+1})$ is infeasible, as the right-hand side of \eqref{eq:cs_obj} depends on the future global model $\x^{\tau+1}$.
\end{problem}
Since $\x^{\tau + 1}$ is explicitly determined by a subset of clients $C_{\tau+1}$, the estimation of $\bE_{C_{\tau+1}} \delta f_\tau(m_{\tau+1})$ requires updates from all workers. Thus, we need to approximate all possible realizations of the future global model $\x^{\tau + 1}$. For this purpose, we propose a \emph{full-client intermediate} communication round $\textstyle \tau+1/2$, i.e. $\textstyle m_{\tau + 1/2} = M$ and $\textstyle C_{\tau+1/2}=\overline{1,M}$. In this round, we will receive client updates $\textstyle \{\x_i^{\tau+1/2}\}_{i=1}^M$, whose combination approximates $\x^{\tau + 1}$ for an arbitrary $C_{\tau+1}$. The suggested intermediate communication leads to the following expectation
\begin{equation}\label{eq:cn_obj_inter}
\begin{aligned}
    & \delta f_{\tau+1/2}(m_{\tau+1}) := \, \bE_{C_{\tau+1}} \delta f_{\tau+1/2}(m_{\tau+1}) = \\
    & = \bE_{C_{\tau+1}}\left[f(\x^{\tau + 1/2}(m_{\tau+1}, C_{\tau+1}))-f(\x^\tau)\right].
\end{aligned}
\end{equation}

\begin{algorithm}[t]
  \caption{\textsc{ParticipantNumberStrategy}}
  \label{alg:number_strategy}
  \begin{algorithmic}[1]
    \Require Iteration $\tau$, previous $m_\tau$, previous model $\x^\tau$.
    \Statex \textbf{Context:} Window $\Delta$, max clients $M$
    \Statex
    \If{$\tau \bmod \Delta \neq 0$}
    \State $m_{\tau+1} \gets m_{\tau}$
    \Else
        \State $\{\x_i^{\tau+1/2}\}_{i=1}^M \gets \textsc{GetUpdates}(\x^\tau, \overline{1,M})$ \label{line:full_update}
        \State \textcolor{blue}{$m_{\tau+1} \gets \textsc{ISP} (\{\x_i^{\tau+1/2}\}_{i=1}^M,  m_\tau)$} \Comment{Algorithm \ref{alg:ISP}}
    \EndIf
    \State \Return $m_{\tau+1}$
  \end{algorithmic}
\end{algorithm}

The current depiction of the federated learning pipeline assumes solution \eqref{eq:select_m} at each communication round (see Line \ref{line:dynamic_number}, Algorithm \ref{alg:fed_periodic}), requiring regular full-client intermediate communications that degrade efficiency.
To resolve it, we pose a dynamic number selection procedure as \emph{strategy}: solve \eqref{eq:select_m} once every $\Delta$ rounds and propagate the resulting $m_{\tau+1}$ to subsequent rounds. We summarize these considerations in Algorithm \ref{alg:number_strategy}. To maintain computational efficiency, clients can participate partially or perform reduced local training epochs during this auxiliary synchronization phase (see Appendix \ref{app:abl:full_comm} for details).

\begin{figure*}[t]
    \centering
    \includegraphics[width=0.95\textwidth]{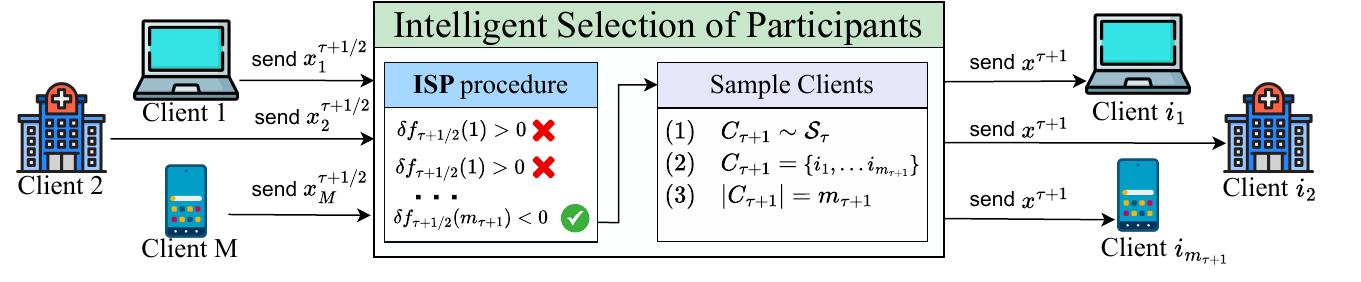}
    \caption{\textsc{ISP} procedure pipeline.}
    \label{fig:pipeline}
\end{figure*}

Intermediate communication with all clients and adaptive selection of their number as a strategy for the next $\Delta$ rounds allows us to formulate an optimization procedure of \eqref{eq:select_m}, which we call \textbf{Intelligent Selection of Participants} (ISP). This workflow is illustrated in Figure \ref{fig:pipeline} and its pseudocode is represented in Algorithm \ref{alg:ISP}. The composition of Algorithms \ref{alg:fed_periodic}-\ref{alg:ISP} defines the overall federated learning pipeline with an adaptive participant count. In Algorithm \ref{alg:ISP}, we loop over counts $m_{\tau+1}$, evaluate the objective in \eqref{eq:cn_obj_inter}, and check its improvement. We now turn to the challenge of estimating this objective.

\begin{problem}[Expectation estimation]\label{problem:expectation_estimation}
    The exact estimation of expectation \eqref{eq:cn_obj_inter} (Line \ref{line:estimate_delta}, Algorithm \ref{alg:ISP}) requires enumerating all $\binom{M}{m_{\tau+1}}$ possible subsets. Thus, the exact solution of \eqref{eq:select_m} is computationally prohibitive.
\end{problem}
To estimate the expected value in \eqref{eq:cn_obj_inter}, we use a \emph{Monte-Carlo} approach, consisting of repetitive sampling $C^n_{\tau+1} \sim \mathcal{S}_\tau$ and aggregating corresponding client updates to $\x^{(\tau+1/2)_n} := \x^{\tau+1/2}(m_{\tau+1}, C^n_{\tau+1})$. The obtained $f(\x^{(\tau+1/2)_n})$ gives an estimate of \eqref{eq:cn_obj_inter} for sufficient coverage
\begin{equation}\label{eq:cn_obj_sample}
\delta f_{\tau+1/2}(m_{\tau+1}) \approx \dfrac{1}{N}\sum_{n=1}^N f(\x^{(\tau+1/2)_n}) - f(\x^{\tau}),
\end{equation}
where $N$ defines the coverage depth and is a hyperparameter of the method (see Appendix \ref{app:abl:num_samples} for its ablation). Algorithm \ref{alg:estimate_expectation} encapsulates this proposition and is a key component of the \textsc{ISP} procedure.

\begin{algorithm}[t]
  \caption{\textsc{ISP} (Intelligent Selection of Participants)}
  \label{alg:ISP}
  \begin{algorithmic}[1]
    \Require Client states $\{\x_i^{\tau+1/2}\}_{i=1}^M$, previous $m_\tau$.
    \Statex \textbf{Context:} max clients $M$, momentum $\beta$, resolution $w$.
    \State $m \gets 1$
    \While{$m \leq M$}
      \State $\delta f_{\tau+1/2}(m) \gets \textsc{ExpectEstim}(\textstyle \{\x_i^{\tau+1/2}\}, m)$ \label{line:estimate_delta}
      \If{$\delta f_{\tau+1/2}(m) < 0$}\,\, \textbf{Break}  \Comment{Early exit} 
      \EndIf
      \State $m \gets m + w$ \label{line:relosution}
    \EndWhile
    \State \Return $\beta m + (1-\beta)m_{\tau}$ \label{line:momentum}
  \end{algorithmic}
\end{algorithm}

\begin{algorithm}[t]
  \caption{\textsc{ExpectEstim}}
  \label{alg:estimate_expectation}
  \begin{algorithmic}[1]
    \Require Client states $\{\x_i^{\tau+1/2}\}_{i=1}^M$, num. participants $m$.
    \Statex \textbf{Context:} strategy $\mathcal{S}_{\tau}$,  previous loss $f(\x^\tau)$, depth $N$.
    \State $f(\x^{\tau+1}) \gets 0$
    \For{$n = 1$ to $N$}
    \State $C^n_{\tau+1} \sim \cc{S}_\tau,\quad |C^n_{\tau+1}| = m$ \Comment{Sample subset}
    \State $\x^{(\tau+1/2)_n} \gets Aggregate(\{\x_i^{\tau+1/2}\}_{i\in C^n_{\tau+1}})$
    \State $f(\x^{(\tau+1)_n}) \gets \textsc{EstimLoss}(\x^{(\tau+1/2)_n}, C^n_{\tau+1})$
    \State $f(\x^{\tau+1}) \gets f(\x^{\tau+1}) + \frac{1}{N}f(\x^{(\tau+1)_n})$
    \EndFor
    \State \resizebox{0.72\linewidth}{!}{$f(\x^{\tau+1}) \gets EMA(\{f(\x^{\tau+1}), f(\x^{\tau}), \ldots\})$} \label{line:smoothing} \Comment{Smoothing}
    \State $\delta f_{\tau+1/2}(m) \gets f(\x^{\tau+1}) - f(\x^\tau)$
    \State \Return $\delta f_{\tau+1/2}(m)$
    \Statex
    \Function{EstimLoss}{$\x, C$}
      \If{$\hat{f}$ available} \label{line:surrogate_aviability}
        \State \Return $\hat{f}(\x)$
    \EndIf
    \ForAll{clients $i \in C$ \textbf{in parallel}}
        \State $f_i(\x) \gets ClientLoss(\x)$
    \EndFor
    \State \Return $\sum_{i\in C} w_i f_i(\x)$
    \EndFunction
  \end{algorithmic}
\end{algorithm}

Despite the naivety of the approximation \eqref{eq:cn_obj_sample}, this approach is not without meaning even in the case of significant client heterogeneity. This is largely due to the moderate variability of the sampling procedure $C^n_{\tau+1} \sim \mathcal{S}_{\tau}$, which is one of the central problems of the client selection strategies. For example, strategies \textsc{Power-of-Choice} \citep{cho2020client} and \textsc{FedCor} \citep{tang2022fedcor} are deterministic, i.e. $N=1$ gives an exact solution of \eqref{eq:select_m}. In \textsc{FedCBS} \citep{zhang2023fed}, the authors apply a sequential sampling strategy to reduce $N$, while in \textsc{DELTA} \citep{wang2023delta} it is directly proven that the proposed unbiased sampling scheme is optimal in terms of minimizing the variance of $\cc{S}_{\tau}$.

\begin{problem}[Greedy optimization]\label{problem:greedy_optimization}
    The greedy nature of optimization \eqref{eq:select_m} contradicts the positioning of defining $m_\tau$ as a $\Delta$-round strategy (see Algorithm \ref{alg:number_strategy}). 
\end{problem}
Despite treating \textsc{ISP} as a strategic procedure, the form \eqref{eq:select_m} still has a greedy nature, since $m_{\tau+1}$ is chosen based on the impact from the previous round state $\x^\tau$. This is exacerbated in the case of client heterogeneity, where the resulting $m_\tau$ may be biased towards a particular round. Thus, we want to have a more conservative procedure \eqref{eq:select_m} that takes into account previous history to capture the global trend in federated learning pipeline. We address this by introducing a $\beta$-\emph{momentum} to updated $m_{\tau+1}$ (see Line \ref{line:momentum}, Algorithm \ref{alg:ISP}) and exponential moving averaging of previous history $f(\x^\tau)$ (see Line \ref{line:smoothing}, Algorithm \ref{alg:estimate_expectation}). For details on their impact, see Appendix \ref{app:abl:momentum}, \ref{app:abl:ema}. 

\begin{problem}[Internal Computation] \label{problem:internal_computation}
    Optimization \eqref{eq:select_m} requires enumerating $m_\tau = \overline{1, M}$. Each enumeration through $m_\tau$ is associated with the computation \eqref{eq:cn_obj_sample}, which requires additional communications with clients.
\end{problem}
To reduce the number of iterations for $m_\tau$, we can introduce a \emph{resolution} factor $w$ for the problem \eqref{eq:select_m}. This factor specifies the step with which we enumerate through $m_\tau$ and reflects the sensitivity of the method to the number of clients in a window of size $2w$ (see Line \ref{line:relosution}, Algorithm \ref{alg:ISP}). For details on its impact, see Appendix \ref{app:abl:resolution}. Computation \eqref{eq:cn_obj_sample} exploits Monte-Carlo technique, which consists of repetitive sampling $C^n_{\tau+1} \sim \mathcal{S}_\tau$ and estimation $f(\x^{(\tau+1/2)_n})$. This procedure requires additional communications with clients and multiple inferences on their devices, which also affects overall performance. Although communication overhead does not pose a significant problem due to the volume of information transmitted from the client to the server, multiple inferences can tax worker devices. This limitation can be addressed by introducing a \emph{surrogate} loss function $\hat f(\x^\tau)$, which is stored on the server and approximates client representations. This is a fairly strong but \emph{optional} assumption (see Line \ref{line:surrogate_aviability}, Algorithm \ref{alg:estimate_expectation}) that makes sense in the case of external publicly available datasets. We demonstrate its applicability in the Experiment \ref{subsec:ECG_exp} on real-world ECG data.
We discuss computational efficiency in cases where the surrogate loss is not available ( Sections \ref{subsec:isp_cs}, \ref{subsec:isp_gc}) in Appendix \ref{app:exp:time}. We also suggest handling such cases in the Appendix \ref{app:abl:full_comm}.

\section{Experiments and Results}\label{sec:experiments_results}

\subsection{Experiment Settings} \label{subsec:exp_settings}

To evaluate the impact of the proposed technique, we conducted extensive experiments on three datasets. The first two are public \textsc{CIFAR-10} \citep{krizhevsky2009learning} and \textsc{Tiny-ImageNet} \citep{le2015tiny}, which represent the multiclass image classification task. The third proprietary dataset (due to privacy issues) consists of approximately 800,000 labeled 12-lead digital ECG segments 10 seconds long, sampled at 500Hz for patients older than 18 years. This dataset was collected in real-world clinical settings, and represents a time series multi-label classification. The independent test set contains approximately 400,000 ECG segments obtained under identical conditions. We utilize \textsc{ResNet-18} \citep{he2016deepresnset} for \textsc{CIFAR-10} and visual transformer \textsc{Swin-T} \citep{liu2021swin} for \textsc{Tiny-ImageNet}. For the ECG domain, we employ an adapted \textsc{ResNet1D-18} with one-dimensional convolutions to handle the multilabel classification objective. The convergence of the model is controlled using the validation part of the local client data.

To demonstrate the capabilities of adaptive selection of the number of clients, all datasets were partitioned among them using the Dirichlet distribution parametrized by the concentration $\alpha$ \citep{hsu2019measuringeffectsnonidenticaldata}. This parameter controls the clients' statistical heterogeneity (strong when $\alpha \to 0$). Table \ref{tab:hyperparameters} summarized the main hyperparameters. More technical and experiments details can be found in the Appendix \ref{app:tech}, \ref{app:exp}.

\begin{table}[t]
\centering
\caption{Hyperparameter setup. CS -- Client Selection (Table \ref{tab:isp_cs_cifar10}), GC -- Gradient Compression (Section \ref{subsec:isp_gc}).
}
\label{tab:hyperparameters}
\resizebox{\columnwidth}{!}{%
\begin{tabular}{
|p{0.32\columnwidth} |
>{\centering\arraybackslash}p{0.44\columnwidth} |
>{\centering\arraybackslash}p{0.24\columnwidth} |
}
\toprule
\textbf{Hyperparameters} & \textbf{Image domain} & \textbf{ECG domain} \\
\midrule
Number of Clients & 100 & 2000 \\
\midrule
Dirichlet \hspace{8mm} concentration & \small\textsc{Tiny-ImageNet}: \hspace{0.3mm} $\alpha = 0.5$ \textsc{CIFAR-10} (CS): \hspace{2mm} $\alpha = 0.1$ \textsc{CIFAR-10} (GC): $\alpha = 1000$ & $\alpha = 5.0$ \\
\midrule
\textsc{ISP}-Solution & $\Delta = 20$ & $\Delta = 20$ \\
\textsc{ISP}-Depth & $N=10$ & $N=20$ \\
\textsc{ISP}-Resolution & $w=1$ & $w=20$ \\
\textsc{ISP}-Momentum & $\beta=0.5$ & $\beta=0.5$ \\
\textsc{ISP}-Surrogate & No & Yes \\
\bottomrule
\end{tabular}%
}
\end{table}

\subsection{Dynamic Participants with Client Selection}\label{subsec:isp_cs}

\paragraph{\textsc{CIFAR-10} Experiments} The \textsc{ISP} methodology explicitly operates with the concept of a client sampling strategy and relies on it in expectation estimation (see Problem \ref{problem:expectation_estimation}), making it natural to assess the impact of an adaptive number of clients on these strategies. 
We consider the baseline strategy of uniform selection of a random subset \citep{mcmahan2017communication}, along with the following state-of-the-art techniques for client sampling mentioned in the related work:

\begin{itemize}
    \item The \textsc{Power-of-Choice} (\textsc{POW-D}) \citep{cho2020client} selects clients by local loss value within a random pool of size $D$. We fix $D=40$ for this ranking;
    \item \textsc{FedCor} \citep{tang2022fedcor}, obtains client losses, approximating them with a Gaussian process; 
    \item \textsc{FedCBS} \citep{zhang2023fed} enforce uniform class-based sampling, balancing exploration vs. exploitation with $\lambda=10$;
    \item \textsc{DELTA} \citep{wang2023delta} in contrast select clients by maximizing external and internal diversities. We set the corresponding proportions $\alpha_1=0.8$ and $\alpha_2=0.2$.
\end{itemize}


The results are presented in Table \ref{tab:isp_cs_cifar10}. We chose the baseline number of clients $m=20$, which is motivated in Appendix, Section \ref{app:exp:m_tau_select}. We observe a significant increase in communication efficiency, up to 25\% for \textsc{FedCor} and \textsc{DELTA} sampling strategies, while maintaining or even slightly improving the quality of the downstream task across all methods. Only for the \textsc{POW-D} baseline, the \textsc{ISP} technique requires barely more communication but demonstrates a substantial increase in the target metrics that we attribute to the adversarial nature of client sampling in \textsc{POW-D}. The communication and accuracy dynamics for this comparison is illustrated in Figure \ref{fig:pow-cifar}. All remaining plots are depicted in Appendix \ref{app:exp:graphic}. We also observe a lower variance in communications for the \textsc{ISP} technique under Uniform sampling. This can be explained by the optimization problem \eqref{eq:select_m}, in which we choose $m_{\tau+1}$ based on the overall model improvement. Despite the approximation \eqref{eq:cn_obj_sample}, $N=10$ was found to be sufficient even in the case of uniform sampling $C^n_{\tau+1}$.
As we discussed in Problem \ref{problem:internal_computation}, the surrogate loss $\hat{f}$ enables additional optimization in expectation estimation (see Line \ref{line:surrogate_aviability}) in terms of client inference. We discuss its impact in Appendix \ref{app:exp:time}. Although its availability is non-trivial in this experiment, we can significantly relax this assumption by reducing the number of participants in the intermediate communication round $m_{\tau+1/2}$ (see Appendix \ref{app:abl:trust}).



\begin{table}[t]
\centering
\caption{Impact of ISP technique with various client selection strategies. Mean and std obtained over 5 runs.}
\label{tab:isp_cs_cifar10}
\resizebox{\columnwidth}{!}{%
\begin{tabular}{
|p{0.23\columnwidth} |
>{\centering\arraybackslash}p{0.3\columnwidth} |
>{\centering\arraybackslash}p{0.25\columnwidth} |
>{\centering\arraybackslash}p{0.25\columnwidth} |
}
\toprule
\textbf{Method} & \textbf{Communications} & \textbf{Test Loss} & \textbf{Accuracy} \\
\midrule
Uniform & 17,767 $\pm$ 1,937 & 0.461 $\pm$ 0.009 & 0.842 $\pm$ 0.004 \\
\textsc{ISP}-Uniform & \textbf{14,343} $\pm$ \hspace*{0.5em} \textbf{556} & 0.464 $\pm$ 0.022 & 0.836 $\pm$ 0.013 \\
\midrule
POW-D & \textbf{10,347} $\pm$ \hspace*{0.5em} \textbf{493}   & 0.573 $\pm$ 0.012 & 0.812 $\pm$ 0.009 \\
\textsc{ISP}-POW-D & 10,864 $\pm$ \hspace*{0.5em} 546    & 0.526 $\pm$ 0.030 & 0.830 $\pm$ 0.011 \\
\midrule
FedCor & 19,360 $\pm$ \hspace*{0.5em} 557   & 0.449 $\pm$ 0.017 & 0.848 $\pm$ 0.006 \\
\textsc{ISP}-FedCor & \textbf{15,037} $\pm$ \hspace*{0.5em} \textbf{468} & 0.439 $\pm$ 0.011 & 0.848 $\pm$ 0.010 \\
\midrule
FedCBS & 19,207 $\pm$ \hspace*{0.5em} 837 & 0.507 $\pm$ 0.018 & 0.830 $\pm$ 0.007 \\
\textsc{ISP}-FedCBS & \textbf{16,611} $\pm$ \hspace*{0.5em} \textbf{816} & 0.487 $\pm$ 0.007 & 0.836 $\pm$ 0.001 \\
\midrule
DELTA & 19,700 $\pm$ \hspace*{0.5em} 191 & 0.816 $\pm$ 0.019 & 0.721 $\pm$ 0.007 \\
\textsc{ISP}-DELTA & \textbf{14,791} $\pm$ \hspace*{0.5em} \textbf{214} & 0.814 $\pm$ 0.029 & 0.722 $\pm$ 0.010 \\
\bottomrule
\end{tabular}%
}
\end{table}

\paragraph{Visual Transformer Experiment} Federated learning often involves complex downstream dependencies. We evaluate classification on Tiny-ImageNet using the Swin-T transformer and apply ISP only with the FedCor sampling strategy to limit computational scaling. Table \ref{tab:isp_imagenet} shows stable scaling and ~11.6\% fewer client–server exchanges alongside marginal performance gains.

\begin{table}[h]
\centering
\caption{\textsc{tiny-ImageNet} setup}
\label{tab:isp_imagenet}
\begin{tabular}{|l|c|c|c|}
\toprule
\textbf{Method} & \textbf{\small Communications} & \textbf{\small Test Loss} & \textbf{\small Accuracy} \\
\midrule
\textsc{\small FedCor} & 1520 & 0.8849 & 0.7945 \\
\textsc{\small ISP FedCor} & \textbf{1344} & \textbf{0.8787} & \textbf{0.7963} \\
\bottomrule
\end{tabular}
\end{table}

\begin{figure}[t]
    \centering
    \includegraphics[width=0.48\textwidth]{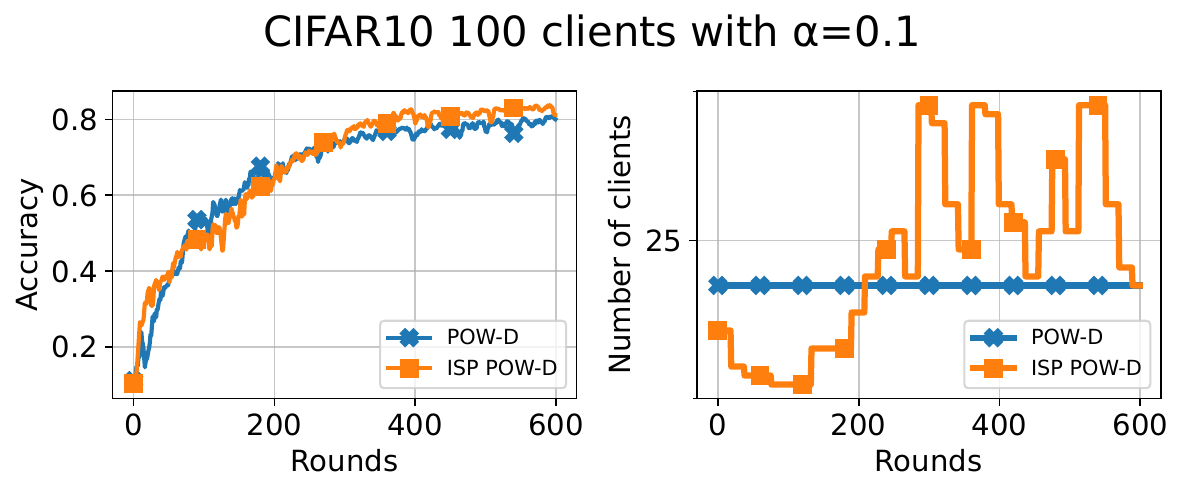}
    \caption{Client count dynamics and test accuracy for \textsc{POW-D} (2nd run). \textsc{ISP-POW-D} achieves optimal performance by round 415, \textsc{POW-D}'s 504. For clarity, results are shown through round 600, omitting $\tau+1/2$ \textsc{ISP} communications.}
    \label{fig:pow-cifar}
\end{figure}










\begin{figure}[t]
    \centering
    \includegraphics[width=0.48\textwidth]{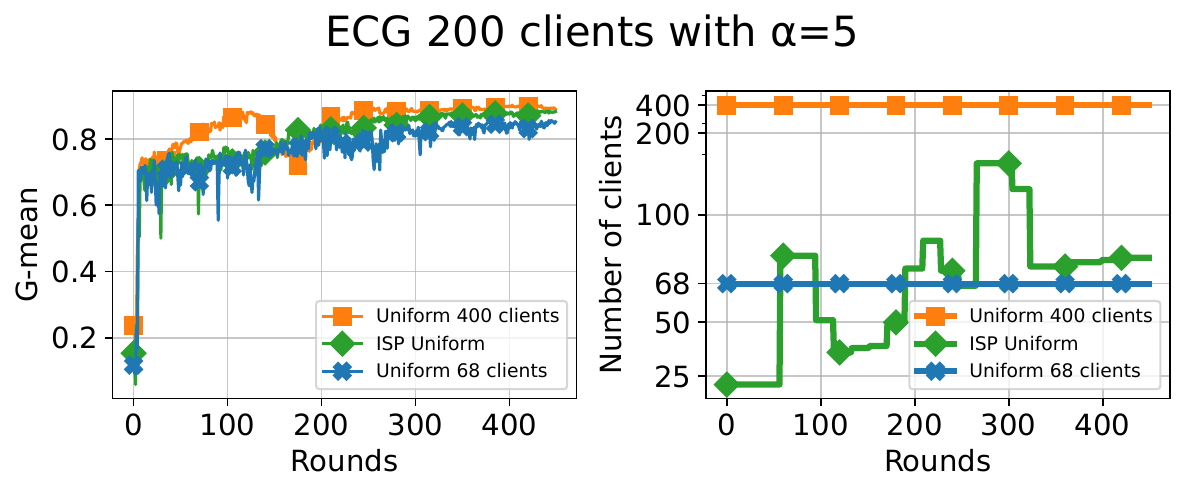}
    \caption{Dynamics of the number of clients and test Geometric mean (G-mean) for \textsc{Uniform} on \textsc{ECG}. \textsc{ISP-POW-D} achieves optimal performance by round 416, \textsc{Uniform} with 400 clients -- 266 round, \textsc{Uniform} with 400 clients -- 420 round. For clarity, results are shown through round 450, omitting $\tau+1/2$ \textsc{ISP} communications.}
    \label{fig:abl_mean_amount_cl_2}
\end{figure}

\subsection{ECG experiment} \label{subsec:ECG_exp}

For ECG classification, we employ a federated learning setup with 2,000 clients, significantly exceeding conventional federated learning scales and providing a rigorous stress-test for our approach. Instead of the absence of the surrogate loss $\hat{f}$ in Section \ref{subsec:isp_cs}, here we derive it from the publicly available PTB-XL dataset \citep{wagner2020ptb}. The client subset size $m = 400$ was chosen following the proportion in Section \ref{subsec:isp_cs}, addressing the non-trivial challenge of optimal client selection in real-world deployments. Successful operation under these conditions validates exceptional robustness and readiness for large-scale industrial applications. 

Since an ECG can be characterized by more than one pathology, we address the multi-label classification of the 6 well-studied \citep{ribeiro2020automatic, hannun2019cardiologist} abnormalities: Atrial FIBrillation (AFIB), Premature Ventricular Complex (PVC), Complete Left (Right) Bundle Branch Block (CLBBB, CRBBB), Left Anterior Fascicular Block (LAFB) and first-degree AV Block (1AVB). A widespread specific of the medical domain is a strong imbalance toward healthy people. To address this, we integrated positive example weights in local client loss functions and used domain-specific metrics such as Sensitivity and Specificity, which are weighted averaged across pathologies \citep{strodthoff2020deep}.

Table~\ref{tab:isp_multilabel} summarizes the experimental results, demonstrating metrics comparable to state-of-the-art ones \citep{ribeiro2020automatic, hannun2019cardiologist}. Our approach reduces communication overhead by more than \textbf{67\%} (compared to \(m=400\)) without substantial degradation in downstream performance. This is largely due to the discrepancy between the median value of the number of clients for the \textsc{ISP} technique and the baseline choice of $m$. These findings highlight not only the method's scalability to large-scale client populations but also its critical practical viability for real-world deployment scenarios when the appropriate number of clients is not known.

\begin{table}[t]
\centering
\caption{Multilabel ECG classification performance comparison ($m=400$). The metrics Spec. (Specificity) and Sens. (Sensitivity) are weighted over target pathologies.}
\label{tab:isp_multilabel}
\resizebox{\columnwidth}{!}{%
\begin{tabular}{|l|c|c|c|}
\toprule
\textbf{Method} & \textbf{Communications} & \textbf{Spec.} & \textbf{Sens.}\\ 
\midrule
Uniform ($m=400$) & 106400 & \textbf{0.9129} & 0.8511\\
\textsc{ISP} Uniform & \textbf{34975} & 0.9049 &  \textbf{0.8596}\\
\bottomrule
\end{tabular}%
}
\end{table}

We quantitatively analyze client selection patterns in our ISP method by examining its average selection rate. Using the clinically realistic ECG task where client participation significantly impacts performance, we observed an average selection of 68 clients per round. To evaluate this behavioral pattern, we compare against a uniform baseline sampling exactly 68 clients. The result of such comparison is illustrated in Figure \ref{fig:abl_mean_amount_cl_2} and in Table \ref{tab:isp_multilabel_m68}. 

While \textsc{ISP} technique has shown a moderate increase in the number of communications, the fixed‐size strategy exhibits a marked degradation in model accuracy, underscoring its inability to adapt to transient fluctuations in client contributions. In contrast, ISP continually reconfigures the participation threshold based on observed gradients, thus mitigating the variance introduced by straggler or low‐utility clients. This comparison reveals that identifying an effective median client count is necessary but not sufficient; only a dynamically adjusted scheme can sustain both communication efficiency and convergence robustness. Consequently, our results demonstrate the indispensable role of temporal adaptivity in participant selection, where ISP not only determines an optimal “median” participation level but also modulates it in response to real‐time performance signals.

\begin{table}[h]
\centering
\caption{Multilabel ECG classification performance comparison ($m=68$). The metrics Spec. (Specificity) and Sens. (Sensitivity) are weighted over target pathologies.}
\label{tab:isp_multilabel_m68}
\resizebox{\columnwidth}{!}{%
\begin{tabular}{|l|c|c|c|}
\toprule
\textbf{Method} & \textbf{Communications} & \textbf{Spec.} & \textbf{Sens.}\\ 
\midrule
Uniform ($m=68$) & \textbf{28560} & 0.8913 & 0.8364\\
\textsc{ISP} Uniform & 34975 & \textbf{0.9049} &  \textbf{0.8596}\\
\bottomrule
\end{tabular}%
}
\end{table}


\subsection{Gradient Compression}\label{subsec:isp_gc} The extensive results on the impact of \textsc{ISP} techniques under client sampling strategies (Tables \ref{tab:isp_cs_cifar10}-\ref{tab:isp_multilabel}) raise reasonable questions about its robustness compared to other approaches to effective communication. Modern architectures are computationally intensive, and despite the progress made with remote devices and the optimization of neural networks, the task of transmitting such a volume of local calculations remains critical and largely determines the communication bottleneck. Therefore, an important communication-efficient practical direction is the compression of transmitted information. We consider the selection of the number of clients in conjunction with \textsc{QSGD} \citep{alistarh2017qsgd} compression. We examine \textsc{TopK} and \textsc{RandK} gradient coordinate selection as compression methods with $K$ equal to $5\%$ and $15\%$ of the total dimension, respectively.
Since compression has difficulty in non-convex convergence for heterogeneous partitioning \citep{sahu2021rethinking}, we conduct uniform ones across clients. The results of the \textsc{ISP} technique along with gradient compression are presented in Table \ref{tab:isp_gc}.

\begin{table}[h]
\centering
\caption{Gradient Compression setup}
\label{tab:isp_gc}
\begin{tabular}{|l|c|c|c|}
\toprule
\textbf{Method} & \textbf{\small Communications} & \textbf{\small Test Loss} & \textbf{\small Accuracy} \\
\toprule
\textsc{\small RandK} & 25040 & 0.4506 & 0.8438 \\
\textsc{\small ISP RandK} & \textbf{17162} & \textbf{0.4447} & \textbf{0.8568} \\
\midrule
\textsc{\small TopK} & 25080 & \textbf{0.4435} & \textbf{0.8650} \\
\textsc{\small ISP TopK} & \textbf{21436} & 0.4644 & 0.8538 \\
\bottomrule
\end{tabular}
\end{table}

It is worth noting that the comparison is also carried out by the number of client-server communications since the volume of transmitted updates is fixed. Thus, the obtained gain demonstrates the principal possibility of integrating \textsc{ISP} technique with different communication-efficient approaches, while achieving up to $30\%$ additional reduction without a significant loss of overall performance. We attribute this increase to the uniform partitioning of the dataset, which is reflected in stricter solutions of \eqref{eq:select_m}. As a consequence, the median number of clients decreases, leading to a more communication-efficient procedure. 



\section{Conclusion}

Our work directly addresses the critical yet underexplored challenge of determining the number of clients per round in federated learning. We systematically approach this problem and demonstrate its relevance, motivating other researchers to pursue developments in this direction. By dynamically optimizing this choice, \textsc{ISP} achieves up to 30\% communication savings while matching or exceeding the accuracy of the leading baselines. Empirical results across diverse tasks underscore the practical significance of client number selection as a standalone problem in FL. These findings highlight its universal impact on efficiency and urge broader attention in the design of federated learning systems. 



\newpage
\bibliography{aaai2026}

\setcounter{secnumdepth}{2}
\newpage
\appendix
\onecolumn
\part*{Appendix}
\renewcommand{\thesection}{\Alph{section}}
\renewcommand{\thesubsection}{\thesection.\arabic{subsection}}
\makeatletter
\renewcommand{\p@subsection}{}
\makeatother
\section{Technical Details}\label{app:tech}

This section provides a comprehensive overview of the technical setup underlying our experiments. We detail the computational resources and software environment used to simulate the federated learning setting, followed by an explanation of key implementation aspects such as client-server orchestration, data partitioning, and communication handling. Finally, we summarize the hyperparameter configurations used across all datasets to ensure fair and consistent comparisons (Table \ref{tab:data_hyperparams}).

\paragraph{Computer resources.} Our implementation is implemented in Python 3.10 and relies on CUDA 12.8 with \textsc{PyTorch} 2.6.0 for GPU acceleration. All experiments (Section \ref{sec:experiments_results}, Appendix \ref{app:exp}, \ref{app:abl}) simulate a distributed setup on a single server, powered by an AMD EPYC 7513 32-core CPU (2.6 GHz base clock) and two NVIDIA A100 SXM4 GPUs (80 GB each) interconnected via NVLink. A detailed description of libraries, versions and code for reproducibility is presented in our codebase \footnote{\url{https://anonymous.4open.science/r/ISPFL/}}.

\paragraph{Code specifics.} The code for our simulation models federated distribution by instantiating a pool of worker processes via Python multiprocessing: each client process is launched with its own pipe endpoint for bidirectional communication with the server and unique rank identifier. A central manager then orchestrates these processes by batching client launches due to the GPU memory constraints of the computer server. All clients are given identical CPU/GPU allocations, and hence the clients are system homogeneous across experiments. 

Within each client, its local dataset is split stratified into training and validation subsets in a ratio of 4 by 1. This division ensures robust global model validation for accurate communication‐cost estimation between experiments with a variable number of clients, and stratification preserves class proportions under heterogeneous Dirichlet sampling (except gradient compression in Section \ref{subsec:isp_gc}). Finally, if any client has fewer than four samples of a given minority class, all such samples are retained in its training partition to avoid splitting inconsistency. We measure communication cost solely by the number of client-to-server exchanges, reflecting the primary bottleneck of transmitting large payloads from distributed nodes. Thus, the \emph{full-intermediate} communication rounds are also taken into account when calculating communications. 

\paragraph{Hyperparameters.} To ensure a fair comparison, we maintain consistent hyperparameters across all methods. Experiment parameters for each dataset are provided in Table \ref{tab:data_hyperparams}, while complete hyperparameter specifications (including normalization factors $\mu_1, \sigma_1, \mu_2, \sigma_2$) are available in our codebase (configs folder).

\begin{table}[h]
\centering
\caption{Parameters for FL experiments across Datasets}
\label{tab:data_hyperparams}
\begin{tabular}{|l|c|c|c|}
\hline
\textbf{Parameter} & \textbf{CIFAR-10} & \textbf{ECG} & \textbf{Tiny ImageNet} \\
\midrule
\textbf{Input Type} & RGB Image & 12-channel Signal & RGB Image \\
\midrule
\textbf{Num Classes} & 10 (Multi-class) & 6 (Multi-label) & 200 (Multi-class) \\
\midrule
\textbf{Model} & ResNet18 & ResNet1D18 & Swin-T \\
\midrule
\textbf{Loss Function} & CrossEntropyLoss & BCEWithLogitsLoss & CrossEntropyLoss \\
\midrule
\textbf{Optimizer} & 
\begin{tabular}{@{}c@{}}Adam \\ SGD (\textsc{DELTA})\end{tabular} &
Adam &
AdamW \\
\midrule
\textbf{Learning Rate} & 
\begin{tabular}{@{}c@{}}Adam: 0.003 \\ SGD (\textsc{DELTA}): 0.003\end{tabular} &
0.003 &
0.001 \\
\midrule
\textbf{Weight Decay} & 
\begin{tabular}{@{}c@{}}Adam: 0 \\ SGD (\textsc{DELTA}): 0\end{tabular} &
0 &
0.01 \\
\midrule
\textbf{Batch Size} & 64 & 128 & 256 \\
\midrule
\textbf{Preprocessing} & $\mathcal{N}(\mu_1, \sigma_1)$ &
$\mathcal{N}(0, 1)$ & $\mathcal{N}(\mu_2, \sigma_2)$ \\
\midrule
\textbf{Augmentations} & 
\begin{tabular}{@{}c@{}}RandomCrop ($32\times32$)\\ HorizontalFlip \end{tabular} & No & No \\
\bottomrule
\end{tabular}
\end{table}

\newpage

\section{Experiments Details}\label{app:exp}

This section presents a thorough account of our experimental evaluation. In Section \ref{app:exp:graphic}, we show additional plots of test loss, accuracy, and client‐count dynamics (see Figures \ref{fig:all_cs_plot}–\ref{fig:comp_plot}). Section \ref{app:exp:m_tau_select} analyzes baseline participant counts under uniform sampling and the resulting communication–quality trade‐offs (Table \ref{tab:choose_m}), while Section \ref{app:exp:time} reports computation‐time measurements for ISP intermediate communications (Table \ref{tab:times} and Table \ref{tab:times_trust}). Finally, Section \ref{app:exp:adafl} compares our adaptive ISP strategy against the linear schedules of AdaFL (Table \ref{tab:isp_adafl}), highlighting the benefits of intelligent client selection.

\subsection{Additional Plots}\label{app:exp:graphic}

\begin{figure}[ht]
    \centering
    \includegraphics[width=0.94\textwidth]{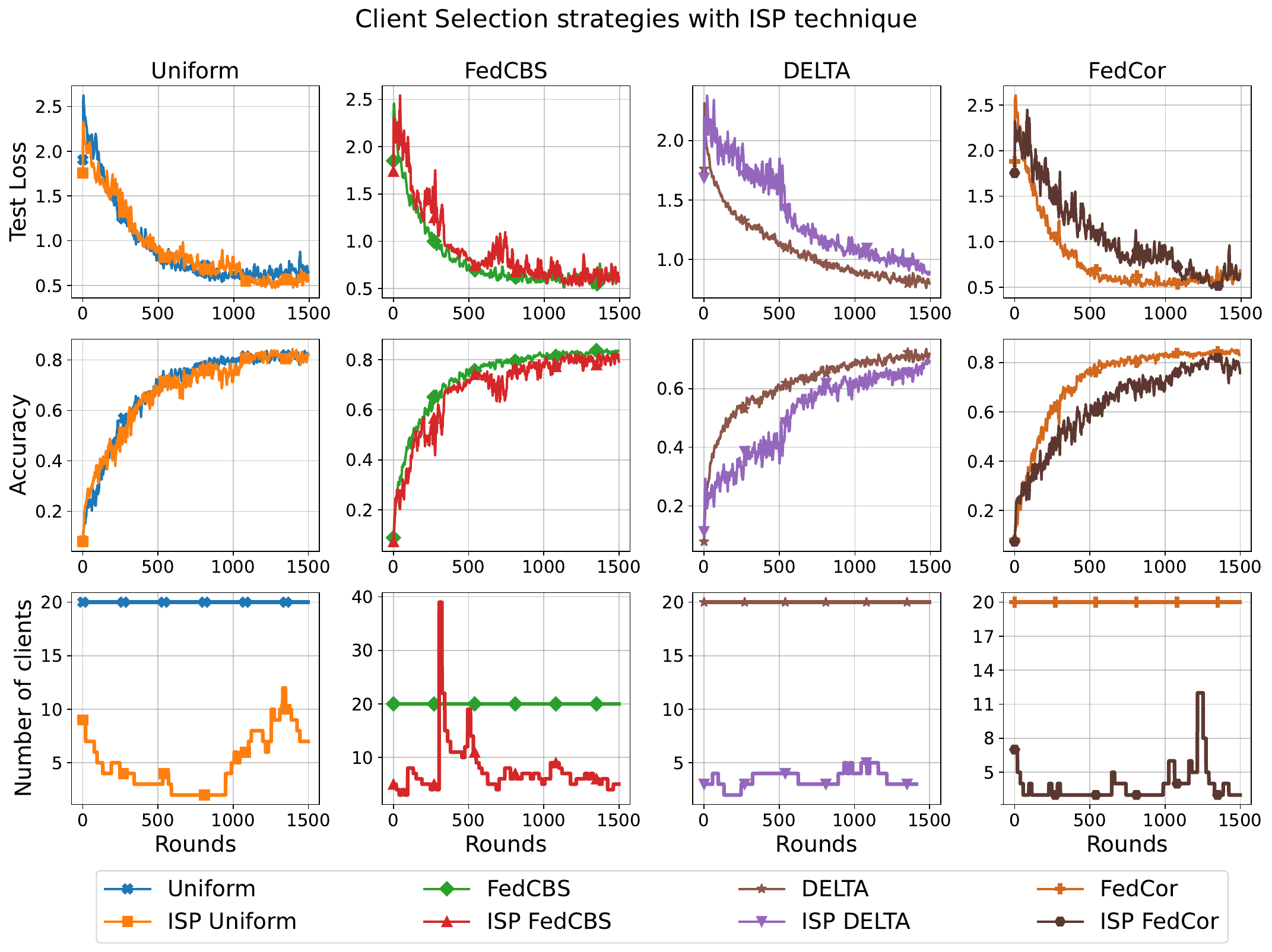}
    \caption{Test loss, accuracy, and client count dynamics, for the \textsc{ISP} technique under various client selection strategies (2nd run). For clarity, results are shown through round 1500, omitting $\tau+1/2$ \textsc{ISP} communications.}
    \label{fig:all_cs_plot}
\end{figure}

\begin{figure}[t]
    \centering
    \includegraphics[width=0.94\textwidth]{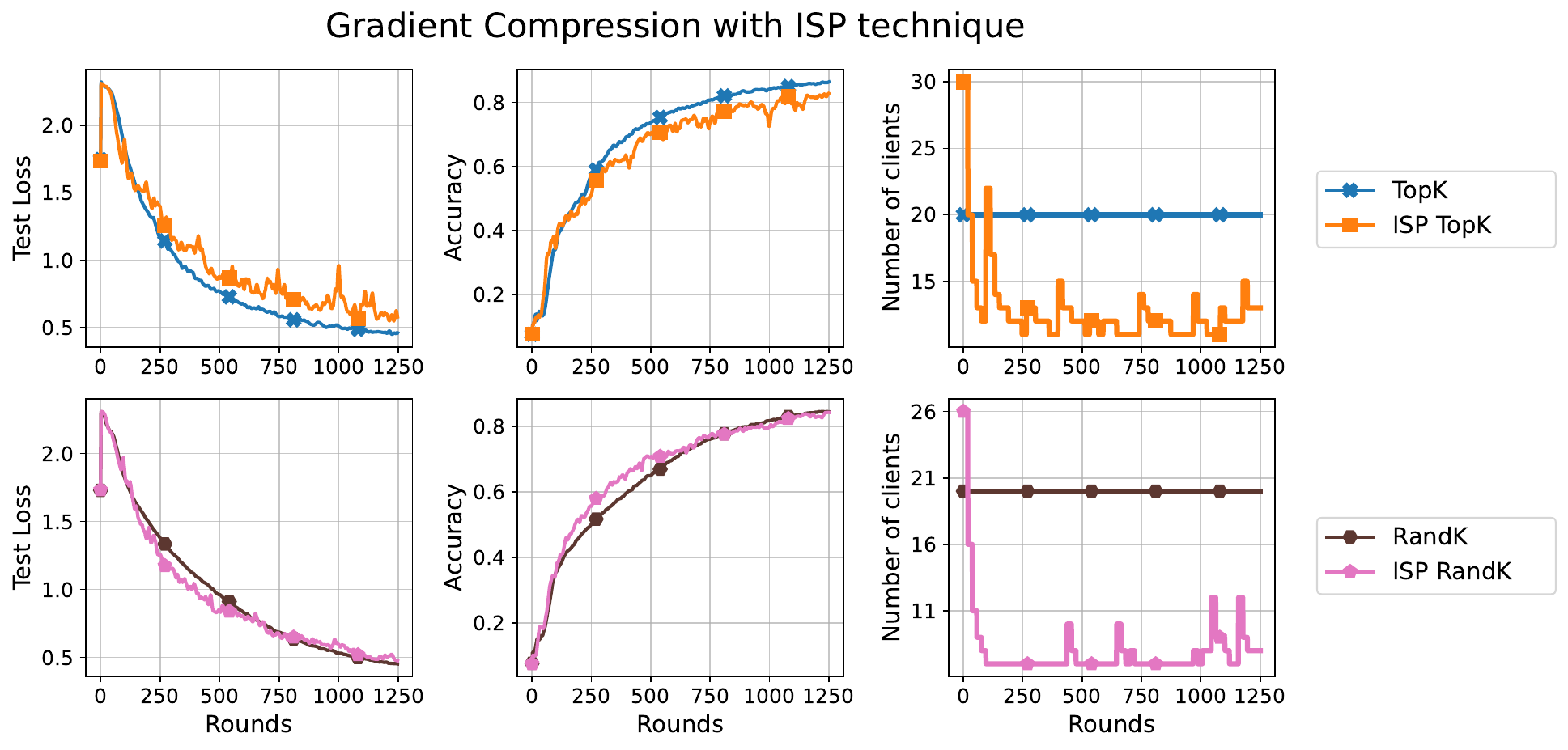}
    \caption{Test loss, accuracy, and client count dynamics, for the \textsc{ISP} technique under various gradient compression methods. We set $K$ equal to $5\%$ and $15\%$ of the total dimension for the \textsc{TopK} and \textsc{RandK} compressions, respectively. For clarity, results are shown through round 1250, omitting $\tau+1/2$ \textsc{ISP} communications.}
    \label{fig:comp_plot}
\end{figure}

In this section, we present and discuss test loss, accuracy, and client count dynamics for the remaining experiments in the main part (Section \ref{sec:experiments_results}). Figures \ref{fig:all_cs_plot}-\ref{fig:comp_plot} illustrate them. Because we performed the experiments five times for Table \ref{tab:isp_cs_cifar10}, the corresponding Figure \ref{fig:all_cs_plot} illustrates the second run, which was selected randomly.

As can be seen from the client count dynamics in Figure \ref{fig:all_cs_plot}, the \textsc{ISP} technique considers between $m=5$ and $m=10$ clients throughout the training for all client strategies except \textsc{FedCBS} (and \textsc{PoW} which we discussed in Section \ref{subsec:isp_cs}). This is consistent with the results of Appendix \ref{app:exp:m_tau_select} (Table \ref{tab:choose_m}) in which the described range of clients leads to communication savings of up to $30\%$ for \textsc{Uniform} sampling. However, in contrast to the baseline with fixed $m$, \textsc{ISP} preserves downstream performance for all client strategies, which is the result of intelligent selection based on the target optimization of \eqref{eq:select_m} and we address them in detail in Appendix \ref{app:exp:m_tau_select}, \ref{app:exp:adafl}. 

This is also reflected in the convergence speed for \textsc{DELTA} and \textsc{FedCor} client strategies, since for them the choice of $m_{\tau+1}$ is closer to $m=5$. Despite the stochastics and slower convergence, the intelligent selection of the number of participants is based on the improvement of the global model, so we observe stable convergence to comparable quality indicators. 

The results in Figure \ref{fig:comp_plot} show the same trends as when adding ISP to client sampling methods, except for the optimal range $m_{\tau+1}$. This again highlights the non-trivial nature of choosing $m$ even for a fixed schedule and demonstrates the practical importance of intelligence selecting the number of participants.

\subsection{Baseline number of participants}\label{app:exp:m_tau_select}

To correctly determine the number of participants $m$ in communication rounds for non-adaptive baselines, we performed an extensive comparison of it for the \textsc{Uniform} client selection strategy under the same conditions as described in Section \ref{subsec:exp_settings} for Table \ref{tab:isp_cs_cifar10}. For $m=20$, we take the average results from the main part of the article (see Table \ref{tab:isp_cs_cifar10}).

\begin{table}[h]
\centering
\caption{The number of participants $m$ for downstream \textsc{CIFAR}-10 task with \textsc{Uniform} client selection strategy.}
\label{tab:choose_m}
\begin{tabular}{
|p{0.08\columnwidth} |
>{\centering\arraybackslash}p{0.15\columnwidth} |
>{\centering\arraybackslash}p{0.1\columnwidth} |
>{\centering\arraybackslash}p{0.1\columnwidth} |
}
\toprule
\textbf{Number of clients} & \textbf{Communications} & \textbf{Test Loss} & \textbf{Accuracy} \\
\midrule
$m=5$  & 12,065 & 0.624 & 0.791 \\
$m=10$ & 15,830 & 0.536 & 0.825 \\
$m=15$ & 18,315 & 0.473 & 0.838 \\
$m=20$ & 17,767 & 0.461 & 0.842 \\
\midrule
$m=25$ & 30,350 & 0.458 & 0.842 \\
$m=40$ & 46,720 & 0.468 & 0.840 \\
$m=50$ & 62,250 & 0.474 & 0.836 \\
\bottomrule
\end{tabular}
\end{table}

Table \ref{tab:choose_m} summarizes this comparison and provides several insights.
\begin{itemize}
    \item \textbf{Diminish returns beyond $m=20$:}  We see a clear non-optimal amount of communication costs for $m > 20$ to obtain the same level of target performance as for $m = 20$. This indicates the well-known statement about the suboptimal nature of a large number of participants in a setup with strong heterogeneity of local client data. 
    \item \textbf{Communication-Quality trade-off:} As we move to $m<20$ we cannot say for sure which of the values $m$ is optimal in terms of communication-quality performance. At $m=10$ we see an acceptable compromise: $11\%$ reduction in communication exchanges while degrading the final accuracy by $1.7\%$. This is especially true in the case where communication efficiency plays a decisive role.
\end{itemize}

Despite the communication efficiency at $m=10$, we preferred to choose $m=20$, focusing on the best final target quality for ease of comparison: while the \textsc{ISP} technique demonstrates a significant reduction in communication costs, downstream performance remains at the same level. In the case of $m=10$, we have the described trade-off, in which excessive communication savings lead to accuracy degradation, which requires additional mention in the main part of the work. Table \ref{tab:isp_cifar_comparison} encapsulates the appropriate numbers of participants $m_\tau$, which is discussed above with an adaptive \textsc{ISP} strategy for comparison. Despite the communication-quality trade-off, \textsc{ISP} demonstrates clear superiority in communication savings and maintains downstream performance, which supports the similar results of Table \ref{tab:isp_multilabel_m68} in the ECG domain.

\begin{table}[t]
\centering
\caption{Comparison of \textsc{ISP} technique against most appropriate number of participants $m$ for \textsc{CIFAR}-10 with \textsc{Uniform} sampling.}
\label{tab:isp_cifar_comparison}
\begin{tabular}{
|p{0.08\columnwidth} |
>{\centering\arraybackslash}p{0.15\columnwidth} |
>{\centering\arraybackslash}p{0.1\columnwidth} |
>{\centering\arraybackslash}p{0.1\columnwidth} |
}
\toprule
\textbf{Number of clients} & \textbf{Communications} & \textbf{Test Loss} & \textbf{Accuracy} \\
\midrule
$m=10$ & 15,830 & 0.536 & 0.825 \\
$m=20$ & 17,767 & \textbf{0.461} & \textbf{0.842} \\
\midrule
\textsc{ISP} $m_\tau$ & \textbf{14,343} & 0.464 & 0.836 \\
\bottomrule
\end{tabular}
\end{table}


\subsection{Computation Time}\label{app:exp:time}

The main motivation for measuring the computation time is to estimate the client inference bottleneck in the absence of a surrogate loss function $\hat{f}$ (see Problem \ref{problem:internal_computation}). To address this challenge in such a setup (see Section \ref{subsec:isp_cs}, Table \ref{tab:isp_cs_cifar10}), we measure the time of the \textsc{ISP} intermediate communication $\textstyle \tau+1/2$ relative to a normal communication round for a baseline sampling strategy. We sampled such communications along the federated learning training pipeline and five runs of repeated experiments. Table \ref{tab:times} encapsulates these measurements.

From the Table \ref{tab:times} we see a clear dependence of the average number of clients selected by the \textsc{ISP} and the relative increase in the time of the intermediate iteration. Thus, the main internal computational complexity of optimization \eqref{eq:select_m} is associated with the enumeration of $m_{\tau+1}$, which we also address in Problem \ref{problem:internal_computation} and Appendix \ref{app:abl:resolution}. For the \textsc{PoW} client strategy, internal computations are also aggravated by the fact that the sampling procedure $C_{\tau+1} \sim \mathcal{S}_\tau$ contains an implicit sorting by local losses from the pool of clients $D$, which requires additional computational costs. We also observer that the intermediate iteration for \textsc{Uniform} client sampling runs on average four times longer than the usual one. However, given that such iterations in this experiment (see Table \ref{tab:isp_cs_cifar10}) were carried out once every $\Delta=20$ rounds, the total loss of efficiency is about $15\%$. Taking into account the optimizations in Appendix \ref{app:abl:delta}, \ref{app:abl:num_samples}, \ref{app:abl:resolution}, which do not lead to communication-quality degradation, such a loss is neutralized to the measurement error.

We further investigate surrogate loss function $\hat f$ implementation to reduce computational overhead and accelerate optimal client count determination. In a supplementary experiment replicating the \textsc{Uniform} strategy setup, we substitute client evaluation with a $\hat f$ as 100-sample dataset subset. Results (Table \ref{tab:times_trust}) demonstrate significant optimization speedup (approximately 2.582$\times$ faster versus full evaluation). Convergence implications are analyzed in Appendix \ref{app:abl:trust}.

This approach eliminates client-side inference during optimization, shifting computational burden to the server while accelerating search procedures. However, reliance on a representative server-side dataset constitutes a strong assumption. We therefore designate surrogate loss implementation as an \textit{optional} methodological enhancement.

\begin{table}[h]
\centering
\caption{Time measurements of \textsc{ISP} intermediate communications in relative to a baseline round. By forming the set across FL pipeline and repeated runs, we calculate the Mean, Std, Max and Min statistics in seconds (s.).}
\label{tab:times}
\begin{tabular}{|l|>{\centering\arraybackslash}p{0.15\textwidth}|c|c|c|}
\toprule
\textbf{Method} & \textbf{Mean relative comparison} & \textbf{Std time solving (s.)} & \textbf{Max time solving (s.)} & \textbf{Min time solving (s.)} \\
\midrule
\textsc{ISP-Uniform} & $\times$3.933  & 36  &  257 & 64 \\
\textsc{ISP-FedCBS} & $\times$7.244  & 72 & 516 & 51 \\
\textsc{ISP-DELTA} & $\times$1.443 & 61 & 396 & 40 \\
\textsc{ISP-POW} & $\times$16.38  & 247 &  1472 & 112 \\
\textsc{ISP-FedCor} &  $\times$1.308  & 18 &  40 & 15 \\
\bottomrule
\end{tabular}
\end{table}

\begin{table}[h]
\centering
\caption{Time measurements of \textsc{ISP} intermediate communications in relative to a baseline round with and without $\hat f$.}
\label{tab:times_trust}
\begin{tabular}{|l|>{\centering\arraybackslash}p{0.15\textwidth}|c|c|c|}
\toprule
\textbf{Method} & \textbf{Mean relative comparison} & \textbf{Std time solving (s.)} & \textbf{Max time solving (s.)} & \textbf{Min time solving (s.)} \\
\midrule
\textsc{ISP-Uniform} & $\times$3.933  & 19  &  168 & 57 \\
\textsc{ISP-Uniform} with $\hat f$ & $\times$1.523  & 6 & 46 & 18 \\
\bottomrule
\end{tabular}
\end{table}

\subsection{AdaFL comparison}\label{app:exp:adafl}
We continue to develop the results of the Table \ref{tab:isp_cifar_comparison} by addressing to the \textsc{ISP} dynamic of $m_\tau$. As we see from Figure \ref{fig:all_cs_plot}, \textsc{ISP-Uniform} has a clear non-linear dynamic with growth of $m_\tau$ at the end, which is consistent with the \textsc{AdaFL} \citep{li2024adafl} mentioned in Related Work (see Section \ref{sec:related_work}). However, the proposed technique represents a linear schedule with a positive trend; as we approach the optimum, we become more sensitive to the heterogeneity of local updates. The original work exploits the schedule from the initial $m_0=10$ to the \emph{full}-round communication $m_T=100$ in a close \textsc{CIFAR-10} setup. However, \textsc{AdaFL} does not directly address communication efficiency and, as we have seen from the results in Table \ref{tab:choose_m}, a significant increase in the number of participants does not provide the same gain in downstream performance that it takes away for communication exchanges. To address the communication bottleneck in addition to the baseline schedule, we will consider a communication-quality trade-off zone ranging clients from $m=5$ to $m=20$ every 50 rounds and then reaching a plateau. We call these schedules \emph{full} and \emph{econom}, respectively. Table \ref{tab:isp_adafl} summarizes the results of the comparison.

\begin{table}[h]
\centering
\caption{Comparison of \textsc{ISP} with linear scheduler dynamics.}
\label{tab:isp_adafl}
\begin{tabular}{|l|c|c|c|}
\toprule
\textbf{Method} & \textbf{Communications} & \textbf{Test Loss} & \textbf{Accuracy} \\
\midrule
\textsc{Uniform} & 17,767 & 0.461 & 0.842 \\
\textsc{AdaFL (econom)} & 16,191  & 0.508 & 0.827 \\
\textsc{AdaFL (full)} & 63,876  & \textbf{0.421} & \textbf{0.855} \\
\midrule
\textsc{ISP-Uniform} & \textbf{14,343} & 0.464 & 0.836 \\
\bottomrule
\end{tabular}
\end{table}

We observe that the linear increasing \emph{econom} scheduler does not achieve the same efficiency as the \textsc{ISP} technique. This demonstrates the need for an intelligent approach to adaptive choice of the number of clients. At the same time, there is a significant spread depending on the chosen schedule scenario and \textsc{AdaFL} does not relieve us from the choice of the boundaries $m_0$ and $m_T$ and the scheduler step, which also significantly affect the federated learning process, as well as the number of clients in the current round.

\newpage

\section{Ablation Study}\label{app:abl}
In this section, we provide an extensive ablation study of the main components of the \textsc{ISP} technique. We discussed them in detail in the main part of Problems \ref{problem:direct_evaluation}-\ref{problem:internal_computation}. The goal of this ablation is to systematically adjust these parameters compared to those chosen in the main part of the experiments (Table \ref{tab:hyperparameters}) to better understand their impact on communication-efficiency caused by the optimization of \eqref{eq:select_m}. By tuning these parameters, we also aim to identify which ones are more or less sensitive (see Appendix \ref{app:abl:full_comm}-\ref{app:abl:ema}). For hyperparameters that directly affect the optimization time \eqref{eq:select_m} (see Appendix \ref{app:abl:num_samples}, \ref{app:abl:resolution}) we perform a comparison of the intermediate round communication times similar to the measurements in Appendix \ref{app:exp:time}. In Appendix \ref{app:abl:optim_problem} we ablate the optimization problem \eqref{eq:select_m} itself, while in Appendix \ref{app:abl:trust} we address the relaxation of the surrogate loss function.

\subsection{Intermediate Participants Number} \label{app:abl:full_comm}

Practical-oriented federated learning scenarios often feature partial client participation due to connectivity or resource constraints \citep{mcmahan2017communication, li2020federated}. Addressing the limitation in Problem \ref{problem:direct_evaluation}, we relax the assumption of universal client availability during intermediate synchronization ($m_{\tau+1/2}=M$). This modification samples subsets of available clients $\textstyle C_{\tau+1/2} \sim \mathcal{S}_\tau$, $\textstyle |C_{\tau+1/2}| < M$ rather than requiring full participation (Line \ref{line:full_update}, Algorithm \ref{alg:number_strategy}). This relaxation immediately limits the approximation \eqref{eq:cn_obj_sample}, since modeling all possible global states of $\textstyle \x^{(\tau+1/2)_n}$ has become infeasible. However, as we discussed in Section \ref{subsec:isp_cs}, even for \textsc{Uniform} sampling, $N=10$ was found to be sufficient to cover $\textstyle \delta f_{\tau+1/2}(m_{\tau + 1})$ with the Monte Carlo approach. This claim is also demonstrated in the Appendix \ref{app:abl:num_samples}, in which we explicitly ablate $N$. These considerations give us optimism that the relaxation of full intermediate communication will not lead to a dramatic degradation of the correct estimation of $m_{\tau+1}$.

\begin{figure}[ht]
    \centering
    \includegraphics[width=0.5\textwidth]{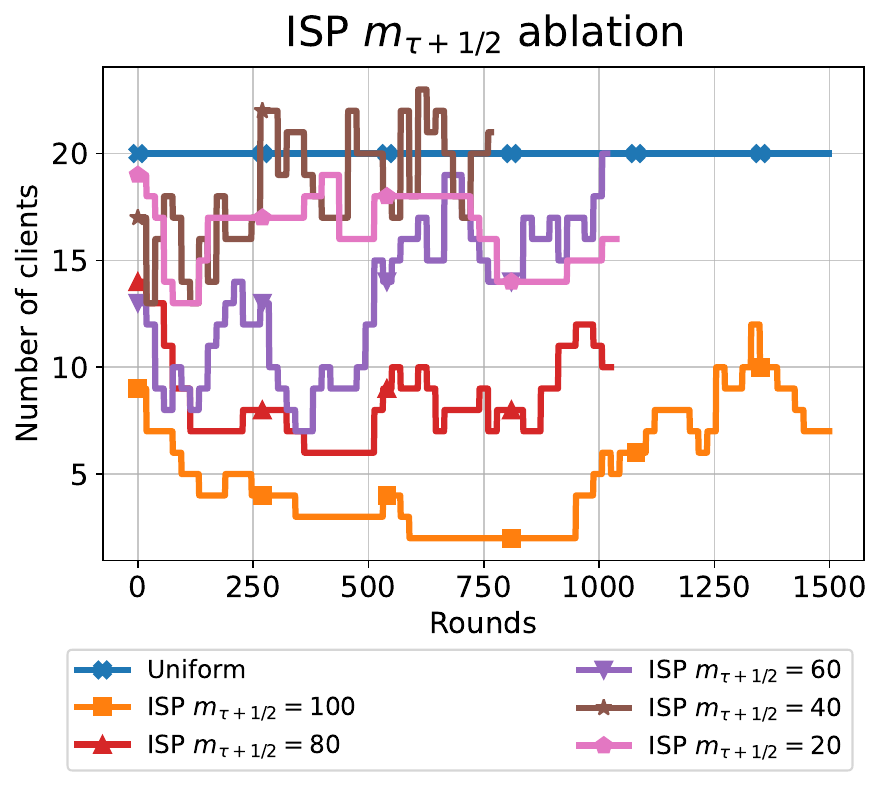}
    \caption{Client count dynamics for the \textsc{ISP} technique under various $m_{\tau+1/2}$ with \textsc{Uniform} sampling strategy. Results are shown up to the best round of each experiment run, as determined by client validation, omitting $\tau+1/2$ \textsc{ISP} communications.}
    \label{fig:abl_full_comm}
\end{figure}

We empirically evaluate this relaxation using the experimental setup shown in Table \ref{tab:isp_cs_cifar10} under the \textsc{Uniform} sampling strategy as the one with the largest variance under statistical heterogeneity of clients for the limiting case of the approximation \eqref{eq:cn_obj_sample}. The results are presented in Table \ref{tab:abl_full_comm} and Figure \ref{fig:abl_full_comm}. From this Figure, we see an interesting dependency: The smaller the number of participants in intermediate communication $m_{\tau+1/2}$ the higher their number $m_{\tau+1}$ participates (except for $m_{\tau+1/2}=20$) in the following $\Delta$ rounds according to the \textsc{ISP} technique. This leads to a predictable increase in communication costs, which in turn increases downstream performance. However, even reducing the full-intermediate round by 5 times ($m_{\tau+1/2}=20$) demonstrates superiority both in communication savings and in the final Accuracy in comparison with the baseline. Interestingly, the choice of $m_{\tau+1/2}=80$ showed a reduction in the number of communications compared to the full round strategy while slightly improving the final quality. We attribute this to the experimental error, which demonstrates the fundamental absence of deterioration of the approximation \eqref{eq:cn_obj_sample} when relaxing the condition to this choice.

This approach also handles Problem \ref{problem:internal_computation} when the surrogate $\hat{f}$ is unavailable. The ability to partially select clients in intermediate communication allows us to set restrictions on them. In terms of the computational efficiency of client inference in the absence of $\hat{f}$, such restrictions can be system or network requirements. This also limits the variability of the resulting $\textstyle \x^{(\tau+1/2)_n}$, which affects the coverage of $\textstyle \delta f_{\tau+1/2}(m_{\tau + 1})$, but this approach allows us to consider the problem of the absence of $\hat{f}$ as one of the restrictions along with the selection of hyperparameters $N$, $\Delta$, and resolution $w$, which is not significant in the practical implementation of \textsc{ISP} technique. We do not provide a relative comparison of the intermediate communication time, since \textsc{ISP-Uniform} did not choose $m_{\tau+1} > 20$ and the ablation of this hyperparameter does not directly affect the amount of computation during optimization \eqref{eq:select_m}.

\begin{table}[h]
\centering
\caption{Intermediate Participants Number $m_{\tau+1/2}$ Ablation.}
\label{tab:abl_full_comm}
\begin{tabular}{|l|c|c|c|}
\hline
\textbf{Method} & \textbf{Communications} & \textbf{Test Loss} & \textbf{Accuracy} \\
\midrule
\textsc{Uniform} & 17,766 & \textbf{0.4605} & 0.8416 \\
\midrule
\textsc{ISP} $m_{\tau+1/2} = 100$ & 14,342 & 0.4635 & 0.8358 \\
\textsc{ISP} $m_{\tau+1/2} = 80$ & \textbf{12,842} & 0.4613 & 0.8403 \\
\textsc{ISP} $m_{\tau+1/2} = 60$ & 15,900 & 0.4666 & 0.8446 \\
\textsc{ISP} $m_{\tau+1/2} = 40$ & 16,873 & 0.4947 & 0.8340 \\
\textsc{ISP} $m_{\tau+1/2} = 20$ & 17,319 & 0.4673 & \textbf{0.8488} \\
\bottomrule
\end{tabular}
\end{table}

\newpage

\subsection{Intermediate Communication Frequency}\label{app:abl:delta}

\begin{figure}[h]
    \centering
    \includegraphics[width=0.5\textwidth]{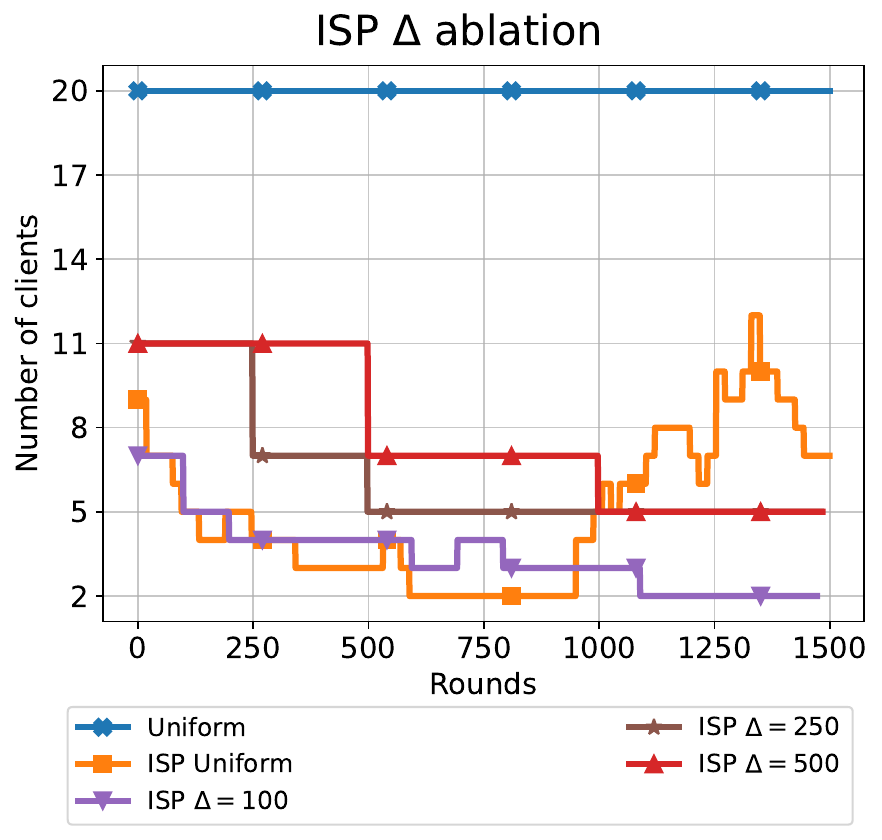}
    \caption{Client count dynamics for the \textsc{ISP} technique under various $\Delta$ with \textsc{Uniform} sampling strategy. For clarity, results are shown through round 1500, omitting $\tau+1/2$ \textsc{ISP} communications.}
    \label{fig:abl_delta}
\end{figure}

Another important hyperparameter of the ISP technique defined in Problem \ref{problem:direct_evaluation} is $\Delta$, which specifies the frequency of solving the problem \eqref{eq:select_m} on the intermediate communication (see Algorithm \ref{alg:number_strategy}). Since collecting updates from all clients incurs substantial computational and communication overhead, minimizing optimization frequency reduces client burden and system costs. This ablation study examines the degradation of downstream performance as $\Delta$ decreases compared to baseline choice of $\Delta=20$ in experimental setup from Section \ref{subsec:isp_cs}. We perform ablation under the \textsc{Uniform} client selection strategy in the same settings as in the Appendix \ref{app:abl:full_comm} for consistency of considerations. Table \ref{tab:abl_delta} and Figure \ref{fig:abl_delta} presents the result.

We see that increasing the frequency $\Delta$ by five times ensured communication savings of up to $12\%$ compared to $\Delta=20$ and up to $28\%$ compared to the baseline, only slightly reducing downstream performance. The subsequent increase in $\Delta$ by another 2.5 times additionally saved $13\%$ of communications with another slight decrease in accuracy. These results demonstrate the compromise Communication-Quality tradeoff described in Appendix \ref{app:exp:m_tau_select} and provide up to $\mathbf{38\%}$ communication efficiency compared to the baseline. However, further rarefaction of the frequency of solving \eqref{eq:select_m} to $\Delta=500$ leads to a significant degradation of performance. We associate this with the blurring of the intelligent dynamics of the adaptive number of participants (see Table \ref{tab:isp_cifar_comparison}, \ref{tab:isp_adafl}), which helps to maintain target performance with communication efficiency due to the type of optimization task \eqref{eq:select_m}. However, the relaxation performed demonstrates a significant gap in the assumptions of this task (see Problems \ref{problem:direct_evaluation}-\ref{problem:internal_computation}) and in the case of $\Delta=250$ with the same Mean Relative comparison (Table \ref{tab:times}) the total computational losses caused by the use of the \textsc{ISP} technique are less than $\mathbf{1.5\%}$.

\begin{table}[h]
\centering
\caption{Intermediate communication frequency $\Delta$ ablation.}
\label{tab:abl_delta}
\begin{tabular}{|l|c|c|c|}
\toprule
\textbf{Method} & \textbf{Communications} & \textbf{Test Loss} & \textbf{Accuracy} \\
\midrule
\textsc{Uniform} & 17,767 & \textbf{0.461} & \textbf{0.842} \\
\textsc{ISP-Uniform} $\Delta = 20$ & 14,343 & 0.464 & 0.836 \\
\midrule
\textsc{ISP-Uniform} $\Delta = 100$ & 12,652 & 0.472 & 0.831 \\
\textsc{ISP-Uniform} $\Delta = 250$ & 10,952 & 0.473 & 0.827 \\
\textsc{ISP-Uniform} $\Delta = 500$ & \textbf{9,602} & 0.480 & 0.818 \\
\bottomrule
\end{tabular}
\end{table}

\newpage
\subsection{Coverage Depth}\label{app:abl:num_samples}
Let us move on to the ablation of the coverage depth $N$, which is defined in Problem \ref{problem:expectation_estimation}. As we noted in the discussion of Table \ref{tab:isp_cs_cifar10}, the baseline choice of $N=10$ may have been excess, since even for the \textsc{Uniform} client sampling strategy, we observed a reduction in variance across 5 experimental runs. This choice could have a negative impact on the final computational performance (see Table \ref{tab:times}), since the coverage depth directly determines the number of client inferences and the complexity of the intermediate communication. The purpose of this ablation is to determine the necessary but sufficient value $N$ of the coverage depth.


\begin{table}[h]
\centering
\caption{Coverage depth $N$ ablation}
\label{tab:abl_num_samples}
\begin{tabular}{|l|c|c|c|}
\toprule
\textbf{Method} & \textbf{Communications} & \textbf{Test Loss} & \textbf{Accuracy} \\
\midrule
\textsc{Uniform} & 17,767 & 0.461 & \textbf{0.842} \\
\textsc{ISP-Uniform} $N = 10$ & 14,343 & 0.464 & 0.836 \\
\midrule
\textsc{ISP-Uniform} $N = 7$ & 14,365 & \textbf{0.453} & \textbf{0.842} \\
\textsc{ISP-Uniform} $N = 5$ & 14,290 & 0.462 & 0.839 \\
\textsc{ISP-Uniform} $N = 3$ & \textbf{14,102} & 0.476 & 0.838 \\
\textsc{ISP-Uniform} $N = 1$ & 15,102 & 0.872 & 0.696 \\
\bottomrule
\end{tabular}
\end{table}

\begin{figure}[ht]
    \centering
    \includegraphics[width=0.5\textwidth]{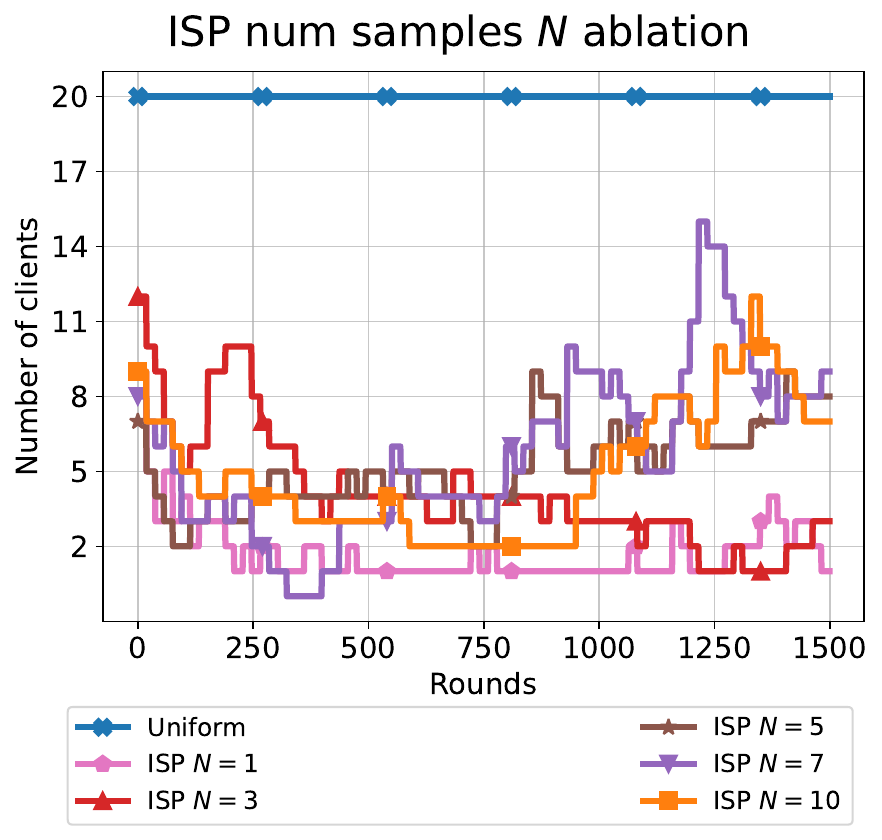}
    \caption{Client count dynamics for the \textsc{ISP} technique under various $N$ with \textsc{Uniform} sampling strategy. For clarity, results are shown through round 1500, omitting $\tau+1/2$ \textsc{ISP} communications.}
    \label{fig:abl_num_samples}
\end{figure}

Table \ref{tab:abl_num_samples} and Figure \ref{fig:abl_num_samples} present the results of the coverage depth ablation. The table confirms the assumption of redundancy of the choice of $N=10$ even in the case of a uniform sampling strategy. We see that a gradual reduction of the coverage depth to $N=3$ does not significantly affect the change in the results. However, for $N=1$ we see a non-obvious behavior, in which it follows from Figure \ref{fig:abl_num_samples} that the \textsc{ISP} technique predominantly selects only one participant, which leads to a dramatic drop in the downstream performance. We address this to expectation estimation (see Algorithm \ref{alg:estimate_expectation}), which relies on the Monte Carlo approach. Then we enumerate through $m_{\tau+1}$ beginning with $m_{\tau+1}=1$, obtain a random $\x^{(\tau+1/2)_i}$, which in the case of one selected participant $i$ is equivalent to the local client update $\x_i^{\tau+1/2}$, and estimate $f(\x_i^{\tau+1/2})$. This update can be unbiased and demonstrate model improvement in terms of $f$, and hence $f(\x_i^{\tau+1/2}) < 0$. Then, according to Algorithm \ref{alg:number_strategy}, we will stop at $m_{\tau+1}=1$, which may not reflect the global needs for FL convergence. In contrast, the choice $N \neq 1$ with high probability under the strong statistical heterogeneity samples the client with a biased update $\x_j^{\tau+1/2}$, which dramatically affects the overall estimate of the left hand side term in \eqref{eq:cn_obj_sample}.

Table \ref{tab:abl_num_samples_time} shows the mean relative time comparison (similar to Table \ref{tab:times}) as $N$ decreases. From Table \ref{tab:abl_num_samples}, we have determined an appropriate sufficient coverage depth $N=3$ that optimizes the number of computations of the solution \eqref{eq:select_m}, making them close to the original round of communication. Thus, with decreasing $\Delta$ frequency, the total computational losses becomes statistically insignificant. This shows a significant gap in Problems \ref{problem:direct_evaluation}-\ref{problem:internal_computation} in practice.

\begin{table}[h]
\centering
\caption{Time measurements of \textsc{ISP} intermediate communications in relative to a baseline round across various coverage depth $N$.}
\label{tab:abl_num_samples_time}
\begin{tabular}{|l|>{\centering\arraybackslash}p{0.15\textwidth}|}
\toprule
\textbf{Method} & \textbf{Mean Relative Comparison} \\
\midrule
\textsc{ISP-Uniform} $N = 10$ & 3.933  \\
\midrule
\textsc{ISP-Uniform} $N = 7$ & 2.753 \\
\textsc{ISP-Uniform} $N = 5$ & 1.966 \\
\textsc{ISP-Uniform} $N = 3$ &  1.180 \\
\textsc{ISP-Uniform} $N = 1$ & 0.393 \\
\bottomrule
\end{tabular}
\end{table}

\subsection{Momentum $\beta$}\label{app:abl:momentum}
The hyperparameter $\beta$ addresses the greedy nature of problem \eqref{eq:select_m} by adjusting the contribution of the resulting $m_{\tau+1}$ compared to the previous history (see Line 8, Algorithm 3). We measure the impact of $\beta$ by disabling it ($\beta=1$) in the same setup setting (see Section \ref{subsec:isp_cs} and Table \ref{tab:hyperparameters}) with the \textsc{Uniform} client strategy. The results are presented in Table \ref{tab:abl_momentum} and the summary Figure \ref{fig:abl_history}, which combines the dynamics of the client count for Appendix \ref{app:abl:momentum} and \ref{app:abl:ema}. 

We see a more stochastic dynamics of the number of selected clients from Figure \ref{fig:abl_history}, which is reflected in a partial loss of communication efficiency without a global improvement in performance (see Table \ref{tab:abl_momentum}). Thus, the $\beta$-momentum in the updated $m_{\tau+1}$ provides greater stability to the greedy optimization in \eqref{eq:select_m}.

\begin{table}[h]
\centering
\caption{Momentum $\beta$ ablation.}
\label{tab:abl_momentum}
\begin{tabular}{|l|c|c|c|}
\toprule
\textbf{Method} & \textbf{Communications} & \textbf{Test Loss} & \textbf{Accuracy} \\
\midrule
\textsc{Uniform} & 17,767 & 0.461 & 0.842 \\
\textsc{ISP Uniform} $\beta=0.5$ & 14,343 & 0.464 & 0.836 \\
\midrule
\textsc{ISP Uniform} $\beta=1$ & 15,786 & 0.462 & 0.834 \\
\bottomrule
\end{tabular}
\end{table}

\newpage
\subsection{Enumerating Resolution}\label{app:abl:resolution}

Another hyperparameter that directly affects the internal computation of optimization \eqref{eq:select_m} (see Problem \ref{problem:internal_computation}) is the resolution $w$, which specifies the enumeration step. This step generates a window $2w$ within which the \textsc{ISP} technique is not sensitive to the selected $m_{\tau+1}$. However, the results of the time measurements in Table \ref{tab:times} demonstrate a direct relationship between the value of $m_{\tau+1}$ and the time of the intermediate communication. Increasing the resolution window clearly allows us to address this challenge.

\begin{figure}[ht]
    \centering
    \includegraphics[width=0.5\textwidth]{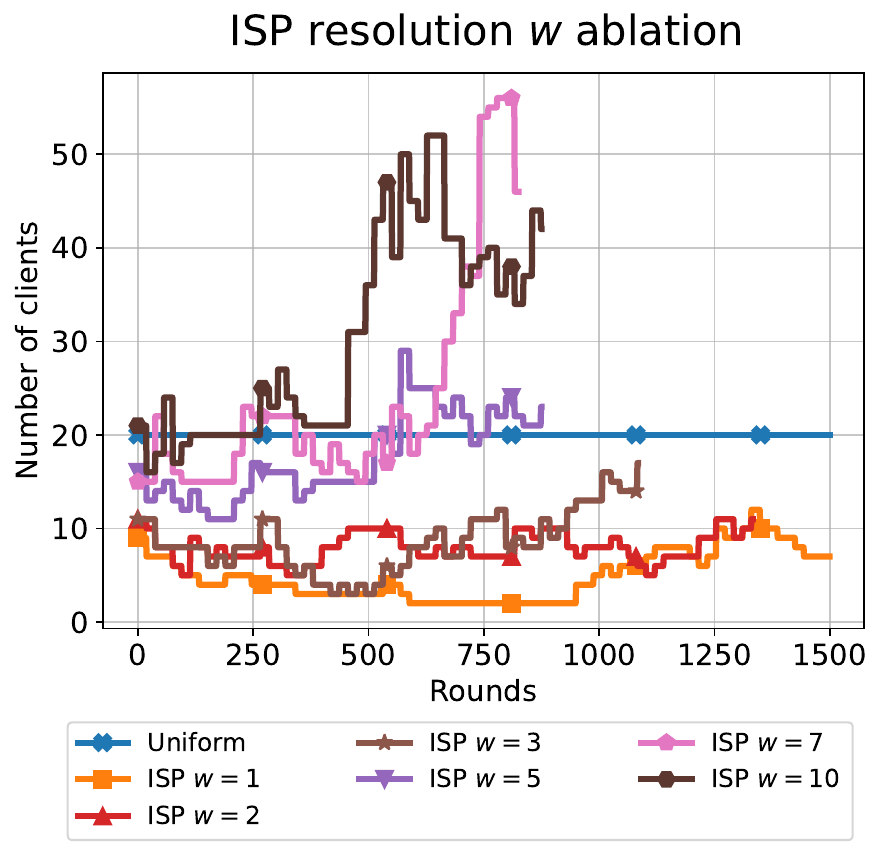}
    \caption{Client count dynamics for the \textsc{ISP} technique under various $N$ with \textsc{Uniform} sampling strategy. Results are shown up to the best round of each experiment run, as determined by client validation, omitting $\tau+1/2$ \textsc{ISP} communications.}
    \label{fig:abl_resolution}
\end{figure}

The results of the ablation are presented in Table \ref{tab:abl_resolution}. Figure \ref{fig:abl_resolution} illustrates the client count dynamics. We see that for $w>3$ the communication costs increase significantly. This is also reflected in the client count dynamics: the average value for $m_{\tau+1}$ increases for $w>3$. We address this to a narrow communication-quality zone from $m = 5$ to $m = 10$ (see the discussion in Appendix \ref{app:exp:m_tau_select}). When we choose a resolution of $w=5$ or more, the \textsc{ISP} technique becomes insensitive to it, which leads to redundant communications at the start, which then generate even more redundant ones due to the greedy structure of the optimized functional \eqref{eq:select_m}. Thus, the resolution $w$ depends on the statistical indicators of the adaptive intelligence dynamics of the participants. However, for $w < 5$ we do not observe significant deviations, while even $w=2$ greatly increases the efficiency of internal computations during optimization of \eqref{eq:select_m}.

We see evidence of the effectiveness of increasing resolution in Table \ref{tab:abl_resolution_time}. For $w=3$, the intermediate communication time becomes comparable to the usual one without a significant decrease in performance. At the same time, a subsequent increase in $w$ does not provide such a noticeable speedup in the computations. This is due to the increase in the selected $m_{\tau+1}$ (see Figure \ref{fig:abl_resolution}), which was discussed in the above paragraph.

\begin{table}[h]
\centering
\caption{Enumerating resolution $w$ ablation.}
\label{tab:abl_resolution}
\begin{tabular}{|l|c|c|c|}
\toprule
\textbf{Method} & \textbf{Communications} & \textbf{Test Loss} & \textbf{Accuracy} \\
\midrule
\textsc{Uniform} & 17,767 & 0.461 & 0.842 \\
\textsc{ISP-Uniform} $w = 1$& 14,343 & 0.464 & 0.836 \\
\textsc{ISP-Uniform} $w = 2$ & 15,771 & 0.467 & 0.842 \\
\textsc{ISP-Uniform} $w = 3$ & 14,515 & 0.483 & 0.837 \\
\textsc{ISP-Uniform} $w = 5$ & 19,739 & 0.467 & 0.852 \\
\textsc{ISP-Uniform} $w = 7$ & 21,939 & 0.464 & 0.856 \\
\textsc{ISP-Uniform} $w = 10$ & 29,984 & 0.458 & 0.860 \\
\bottomrule
\end{tabular}
\end{table}

\begin{table}[h]
\centering
\caption{Time measurements of \textsc{ISP} intermediate communications in relative to a baseline round across various resolutions $w$.}
\label{tab:abl_resolution_time}
\begin{tabular}{|l|>{\centering\arraybackslash}p{0.15\textwidth}|}
\toprule
\textbf{Method} & \textbf{Mean Relative Comparison} \\
\midrule
\textsc{ISP-Uniform} $w = 1$ & 3.933  \\
\midrule
\textsc{ISP-Uniform} $w = 2$ & 1.814 \\
\textsc{ISP-Uniform} $w = 3$ & 1.206 \\
\textsc{ISP-Uniform} $w = 5$ &  1.068 \\
\textsc{ISP-Uniform} $w = 7$ &  1.011 \\
\textsc{ISP-Uniform} $w = 10$ &  1.137 \\
\bottomrule
\end{tabular}
\end{table}

\newpage
\subsection{Exponential Moving Average}\label{app:abl:ema}

\begin{figure}[ht]
    \centering
    \includegraphics[width=0.38\textheight]{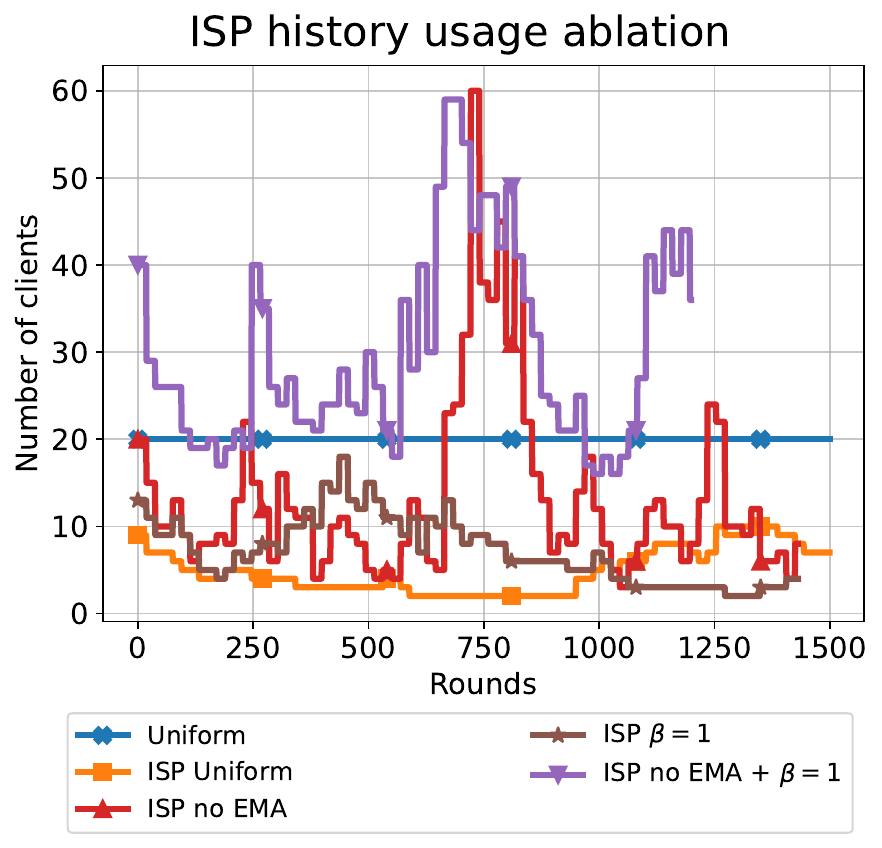}
    \caption{Client count dynamics for the \textsc{ISP} technique with various history usage technique with \textsc{Uniform} sampling strategy. Results are shown up to the best round of each experiment run, as determined by customer validation, omitting $\tau+1/2$ \textsc{ISP} communications.}
    \label{fig:abl_history}
\end{figure}

As discussed in Problem \ref{problem:greedy_optimization}, we employ Exponential Moving Average (EMA) with window size $w_{ema}=5$ to smooth global loss histories. Figure \ref{fig:all_cs_plot} shows noticeable noise in the test loss due to significant statistical heterogeneity in local clients data. This can potentially compromise loss-based optimization \eqref{eq:select_m}. This study ablates EMA's impact on the \textsc{ISP} procedure. Table \ref{tab:abl_ema} and Figure \ref{fig:abl_history} encapsulate the overall results of history ablation (Appendices \ref{app:abl:momentum}, \ref{app:abl:ema}). 

Without EMA smoothing, we observe more unstable client selection patterns (see Figure \ref{fig:abl_history}). This stochasticity increases communication costs beyond the \textsc{Uniform} baseline despite comparable convergence. To measure the impact of Problem \ref{problem:greedy_optimization} suggestions, we conduct an additional experiment that disables both EMA smoothing and participant momentum ($\beta = 1$), eliminating historical context.
The obtained results continue the intuition about the importance of the previous context: we see a more unstable dynamic of client selection, which further leads to communication overhead.

\begin{table}[h]
\centering
\caption{History usage ablation.}
\label{tab:abl_ema}
\begin{tabular}{|l|c|c|c|}
\toprule
\textbf{Method} & \textbf{Communications} & \textbf{Test Loss} & \textbf{Accuracy} \\
\midrule
\textsc{Uniform} & 17,767 & 0.461 & 0.842 \\
\textsc{ISP Uniform} & 14,343 & 0.464 & 0.836 \\
\textsc{ISP Uniform no EMA} & 24,445 & 0.466 & 0.839 \\
\textsc{ISP Uniform no EMA, $\beta = 1$} & 26,905 & 0.469 & 0.822 \\
\bottomrule
\end{tabular}
\end{table}

\newpage
\subsection{ISP Relative Improvement}\label{app:abl:optim_problem}

We revisit the formulation of an intelligent selection of participants \eqref{eq:select_m}, to emphasize a \emph{relative} communication costs. The communication efficiency nature of optimization \eqref{eq:select_m} prioritizes communication reduction by selecting a minimal number $m_{\tau+1}$ of clients that provide positive convergence contributions. Although this approach is effective for overhead minimization, it may sacrifice global convergence optimum. We therefore explore a different adaptive number strategy by explicitly modeling the Quality-Communication trade-off through \textsc{ISP} Relative Improvement (\textsc{ISP-RI}) optimization:

\begin{equation} \label{eq:isp_relative_improve}
m_{\tau + 1} = \arg \max_{m} \quad \frac{\bE_{C_{\tau+1}} \delta f_\tau(m)}{m^\alpha},
\end{equation}

where $\alpha > 0$ is a hyperparameter that determines the relationship in communication-quality trade-off. We set $\alpha = 1$ in the experiment below. This objective function maximizes improvement-per-client rather than merely ensuring positive contributions, naturally shifting focus toward model quality. The implementation of intelligent selection naturally inherits \textsc{ISP} pipeline (Algorithm \ref{alg:number_strategy}- \ref{alg:estimate_expectation}), since the optimization procedure faces the same Problems \ref{problem:direct_evaluation}-\ref{problem:internal_computation} from the original formulation. Note that the iteration loop in Algorithm \ref{alg:ISP} will be different: optimizing the Relative Improvement requires a complete enumeration of existing $m=\overline{1, M}$, which does not allow for an early exit. Without surrogate loss $\hat f$, clients need to participate in these inferences which can lead to their burden. Future work aims to ablate the complexity reduction techniques, particularly leveraging our method's hyperparameters: Communication frequency $\Delta$ (see Appendix \ref{app:abl:delta}), coverage depth $N$ (see Appendix \ref{app:abl:num_samples}), and enumerating resolution $w$ (see Appendix \ref{app:abl:resolution}).

\begin{table}[h]
\centering
\caption{Comparison of \textsc{ISP-RI} with \textsc{Uniform} client sampling strategy}
\label{tab:isp_ri}
\begin{tabular}{|l|c|c|c|}
\toprule
\textbf{Method} & \textbf{Communications} & \textbf{Test Loss} & \textbf{Accuracy} \\
\midrule
\textsc{Uniform} & 17,767 & 0.461 & 0.842 \\
\textsc{AdaFL (econom)} & 16,191  & 0.508 & 0.827 \\
\textsc{AdaFL (full)} & 63,876  & 0.421 & 0.855 \\
\midrule
\textsc{ISP Uniform} & \textbf{14,343} & 0.464 & 0.836 \\
\textsc{ISP-RI Uniform} & 49,594 & \textbf{0.397} & \textbf{0.873} \\
\bottomrule
\end{tabular}
\end{table}

\begin{figure}[ht]
    \centering
    \includegraphics[width=0.96\textwidth]{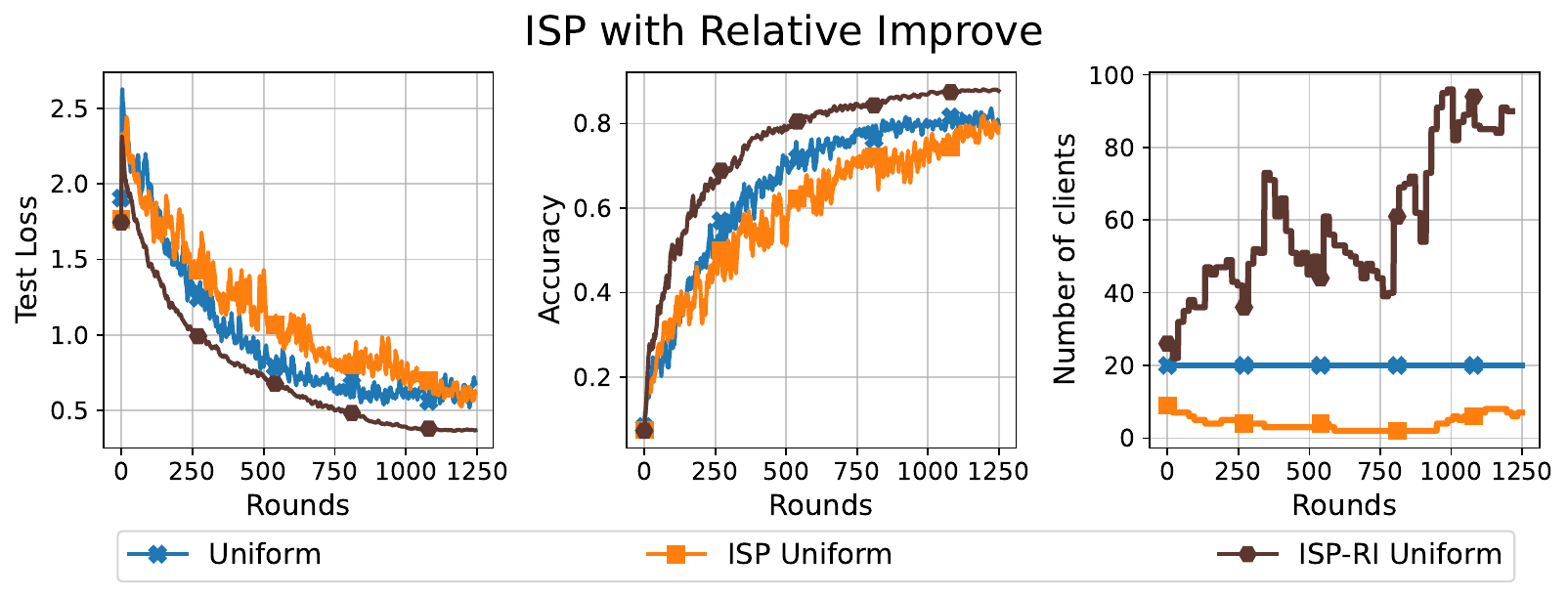}
    \caption{Test loss, accuracy, and client count dynamics, for the \textsc{ISP} and \textsc{ISP-RI} technique with \textsc{Uniform} client selection strategy. For clarity, results are shown through round 1250, omitting $\tau+1/2$ \textsc{ISP} communications.}
    \label{fig:abl_task}
\end{figure}

We compare \textsc{ISP-RI} with baseline \textsc{Uniform} client sampling, \textsc{ISP} strategy with functional optimization \eqref{eq:select_m}, and also with linear increase of \textsc{AdaFL}, since the \emph{full}-schedule (see Appendix \ref{app:exp:adafl}) addresses the quality trade-off with communication overhead. Table \ref{tab:isp_ri} summarizes the results of this comparison. Figure \ref{fig:abl_task}) illustrates the convergence dynamic. The \textsc{ISP-RI} approach achieves superiority over all above methods in terms of downstream quality and convergence speed. It also has a positive dynamics of the number of clients as the full-schedule of \textsc{AdaFL}. At the same time, in contrast to the linearity of the baseline schedule, we see a clear nonlinear dependence. As a result, we observe a significant increase in downstream performance with communication savings up to 12\%. Furthermore, \textsc{ISP-RI} maintains a more stable convergence. However, \textsc{ISP-RI} is significantly inferior to optimization \eqref{eq:select_m} in terms of communications savings, which reflects the communication-quality trade-off that we discussed in Appendix \ref{app:exp:m_tau_select}.

Fundamentally, \textsc{ISP-RI} complements the existing results and demonstrates a flexible and extensive mechanism of Our Methodology (see Section \ref{subsec:our_methodology}) to the scenarios where attention shifts from communication costs to convergence is critical.

\newpage
\subsection{Relaxing trust assumption}\label{app:abl:trust}

Building upon our client computation reduction strategies mentioned in Section \ref{subsec:isp_cs} (specifically partial participation where $m_{\tau + 1/2} < M$, analyzed in Appendix \ref{app:abl:full_comm}), we now examine an orthogonal approach: relaxing requirements for the server-side surrogate loss $\hat f$ introduced in Problem \ref{problem:internal_computation} and Appendix \ref{app:exp:time}. While $\hat f$ effectively transfers computational burden from clients to the server and accelerates optimization, its assumed availability represents a strong practical constraint. This ablation study quantifies our method's sensitivity to $\hat f$ quality, specifically investigating whether performance degradation occurs when using non-ideal surrogate loss functions.

Using the \textsc{Uniform} strategy setup from Table \ref{tab:isp_cs_cifar10}, we implement two configurations:
\textsc{Normal $\hat f$}: 100-example training subset and
\textsc{Corrupted $\hat f$}: Same subset with aggressive augmentations (for a detailed description of augmentations, see Table \ref{tab:abl_trust_aug_params}).

\begin{figure}[ht]
    \centering
    \includegraphics[width=0.5\textwidth]{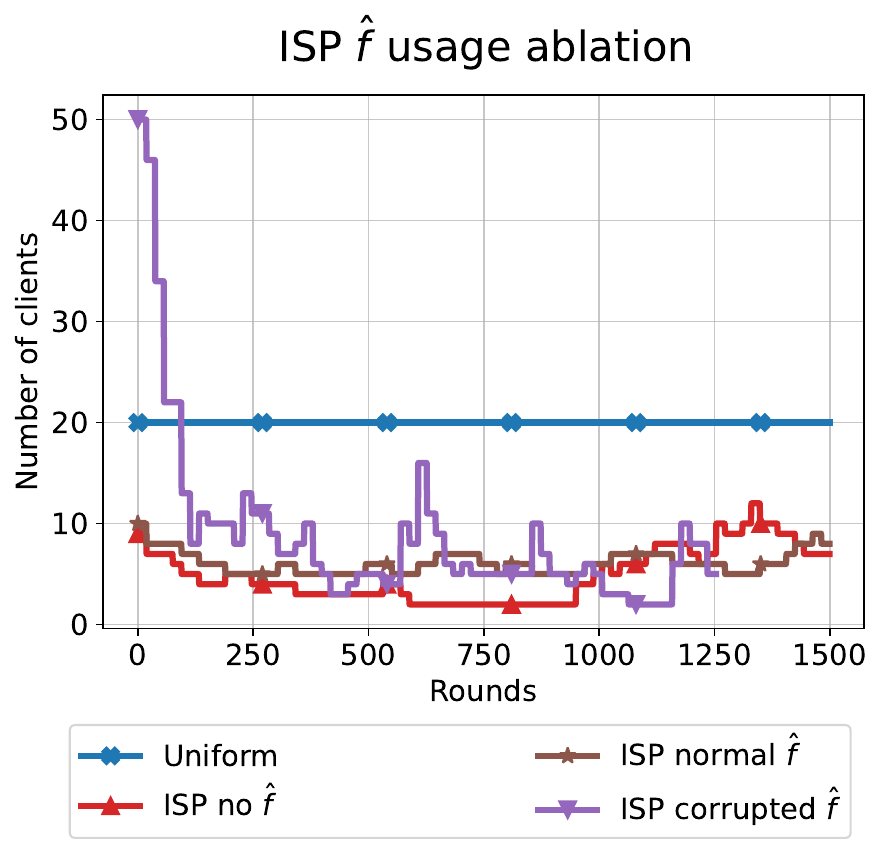}
    \caption{Client count dynamics for the \textsc{ISP} technique with various $\hat f$ surrogate loss usage with \textsc{Uniform} sampling strategy. For clarity, results are shown through round 1500, omitting $\tau+1/2$ \textsc{ISP} communications.}
    \label{fig:abl_trust}
\end{figure}

Results (Table \ref{tab:abl_trust}, Figure \ref{fig:abl_trust}) demonstrate remarkable consistency: performance metrics and client selection dynamics curves remain nearly identical across standard, corrupted, and $\hat f$-free configurations. This minimal sensitivity to $\hat f$ quality indicates our method maintains operational integrity even with substantially degraded surrogate $\hat f$.

\begin{table}[h]
\centering
\caption{Augmentation parameters for corrupted $\hat f$}
\label{tab:abl_trust_aug_params}
\renewcommand{\arraystretch}{0.9} 
\begin{tabular}{|l|c|>{\centering\arraybackslash}m{9.5cm}|}
\toprule
\textbf{Augmentation Technique} & \textbf{Parameters} & \textbf{Description} \\ 
\midrule
\addlinespace
\textsc{RandomGrayScale} & 
\begin{tabular}{@{}c@{}} p=0.8 \end{tabular} & 
Randomly converts images to grayscale with high probability. Significantly reduces color information. \\
\addlinespace
\hline
\addlinespace

\textsc{ColorJitter} & 
\begin{tabular}{@{}c@{}} brightness=0.5 \\ contrast=0.5 \\ saturation=0.5 \\ hue=0.3 \end{tabular} & 
Applies strong random color distortions. Severely alters color distribution and visual appearance. \\
\addlinespace
\hline
\addlinespace

\textsc{RandomResizedCrop} & 
\begin{tabular}{@{}c@{}} size=32 \\ scale=(0.5, 1.0) \end{tabular} & 
Crops random portions of image and resizes. May remove critical features and distort object shapes. \\
\addlinespace
\hline
\addlinespace

\textsc{RandomHorizontalFlip} & 
\begin{tabular}{@{}c@{}} p=0.5 \end{tabular} & 
Flips image horizontally with 50\% probability. Preserves content but alters spatial orientation. \\
\addlinespace
\hline
\addlinespace

\textsc{RandomVerticalFlip} & 
\begin{tabular}{@{}c@{}} p=0.3 \end{tabular} & 
Flips image vertically with 30\% probability. Dramatically changes scene orientation (uncommon in natural images). \\
\addlinespace
\hline
\addlinespace

\textsc{GaussianBlur} & 
\begin{tabular}{@{}c@{}} kernel size=3 \\ $\sigma$=(0.1, 2.0) \end{tabular} & 
Adds variable blurring effect. Reduces high-frequency details and sharpness. \\
\addlinespace
\hline
\addlinespace

\textsc{Normalization} & 
\begin{tabular}{@{}c@{}} $\mu$=(0.485, 0.456, 0.406) \\ $\sigma$=(0.229, 0.224, 0.225) \end{tabular} & 
Standard channel-wise normalization. Transforms pixel distributions but preserves visual content. \\
\addlinespace
\bottomrule
\end{tabular}
\end{table}

\begin{table}[h]
\centering
\caption{Surrogate loss $\hat f$ ablation.}
\label{tab:abl_trust}
\begin{tabular}{|l|c|c|c|}
\toprule
\textbf{Method} & \textbf{Communications} & \textbf{Test Loss} & \textbf{Accuracy} \\
\midrule
\textsc{Uniform} & 17,767 & 0.461 & 0.842 \\
\textsc{ISP no $\hat f$} & 14,343 & 0.464 & 0.836 \\
\textsc{ISP normal $\hat f$} & 14,525 & 0.462 & 0.837 \\
\textsc{ISP corrupted $\hat f$} & 14,740 & 0.466 & 0.849 \\
\bottomrule
\end{tabular}
\end{table}

\end{document}